\documentclass[preprint,12pt]{elsarticle}

\usepackage{amssymb}
\usepackage{amsmath}
\usepackage[dvipsnames]{xcolor}
\usepackage{caption}
\usepackage{subcaption}
\usepackage{float}
\usepackage{algorithm}
\usepackage{algorithmic}
\journal{Neural Networks}

\begin{document}

\begin{frontmatter}



\title{Regularized Reward-Punishment Reinforcement Learning} 


\author[label1]{Jiexin Wang} 

\affiliation[label1]{organization={Dept. of Brain Robot Interface, ATR Computational Neuroscience Laboratories},
            addressline={2-2-2}, 
            city={Hikaridai, Seikacho, Soraku-gun, Kyoto},
            postcode={619-0288}, 
            country={Japan}}

\author[label1]{Eiji Uchibe} 

\begin{abstract}
We propose KL-Coupled Policy Regularization (KCPR), a policy coordination framework for Reward–Punishment Reinforcement Learning (RPRL). Based on KCPR, we derive KL-Coupled Soft Optimality (KCSO) and develop its deep realization, klDMP. Unlike existing RPRL approaches that optimize reward-seeking and punishment-related policies largely independently, KCPR enables direct interactions between companion policies by treating each as a dynamically learned prior for the other. KCSO yields coupled soft-optimal policies and KL-regularized Bellman operators, allowing reward and punishment information to jointly influence value propagation. To improve learning stability, we introduce a companion-prior softening mechanism and evaluate separate replay-buffer designs for balancing reward- and punishment-related experience. Experiments in grid-world and Gazebo robotic navigation tasks demonstrate that klDMP improves safety and learning stability while maintaining competitive task performance compared with DQN, SQL and softDMP. These results suggest that policy-level coordination provides an effective mechanism for integrating multiple behavioral objectives and may serve as a useful design principle for reinforcement learning systems with interacting motivational processes.
\end{abstract}


\end{frontmatter}



\section{Introduction}
\label{sec:intro}
Intelligent agents operating in the real world must continuously negotiate between multiple reward-related signals, including the pursuit of desirable outcomes and the avoidance of harmful consequences under uncertainty. While standard Reinforcement Learning (RL) typically models this trade-off using a single scalar reward, biological decision-making systems rely on distributed and interacting motivational circuits rather than a unidimensional signal. Neuroscience studies suggest that reward and punishment are processed through distinct but interacting pathways, shaped by different neuromodulatory mechanisms and contributing jointly to adaptive behavior \citep{Seymour2007a,Seymour2012a,Eldar2016a}. Importantly, these systems do not operate independently—they influence each other’s learning dynamics in structured ways that support robust decision-making.

This biological view has motivated the Reward–Punishment Reinforcement Learning (RPRL) paradigm, where positive and negative outcomes are processed through separate computational modules. Frameworks such as MaxPain \citep{Elfwing2017a}, Deep MaxPain (DMP) \citep{Wang2017b, Wang2021a} and softDMP \citep{Wang2024} demonstrate that decomposed value systems can accelerate early learning, induce safe exploration and improve sample efficiency and robustness, particularly in robot navigation tasks where adverse outcomes like collisions are essential learning signals. Yet existing RPRL architectures maintain an important simplifying assumption: goal-seeking and risk-aware behaviors develop independently, interacting only through value mixing or uniform-prior entropy. This assumption limits the expressive capability of decomposed RL systems, preventing the two pathways from shaping each other’s behavior in an evolving and structured manner seen in biological agents.

Our key insight is that interactions between motivational systems should occur at the policy level rather than solely through value decomposition. To this end, we introduce KL-Coupled Policy Regularization (KCPR), a principle in which each policy is regularized by a dynamically learned companion policy. 
Unlike conventional KL regularization, which is typically used for soft-optimality and entropy regularization \citep{Azar2012a,Fox2016a,Haarnoja2017a,Haarnoja2018a,Wang2024} or trust-region policy optimization \citep{peters2010relative,schulman2015trust,schulman2017proximal}, KCPR uses KL divergence as a mechanism for coordinating multiple motivational systems.
Based on KCPR, we derive KL-Coupled Soft Optimality (KCSO), a soft-optimality framework in which companion policies act as adaptive priors, yielding coupled soft-optimal policies and Bellman operators that allow reward and punishment information to influence each other's value propagation.
We further develop klDMP as a practical realization of KCSO within the RPRL paradigm. To improve learning stability and data efficiency, we introduce a softening mechanism for companion priors and revisit the separate replay-buffer scheme proposed in softDMP \citep{Wang2024} under the KL-coupled setting.

\begin{figure}[H]
\centering
\includegraphics[width=0.8\hsize]{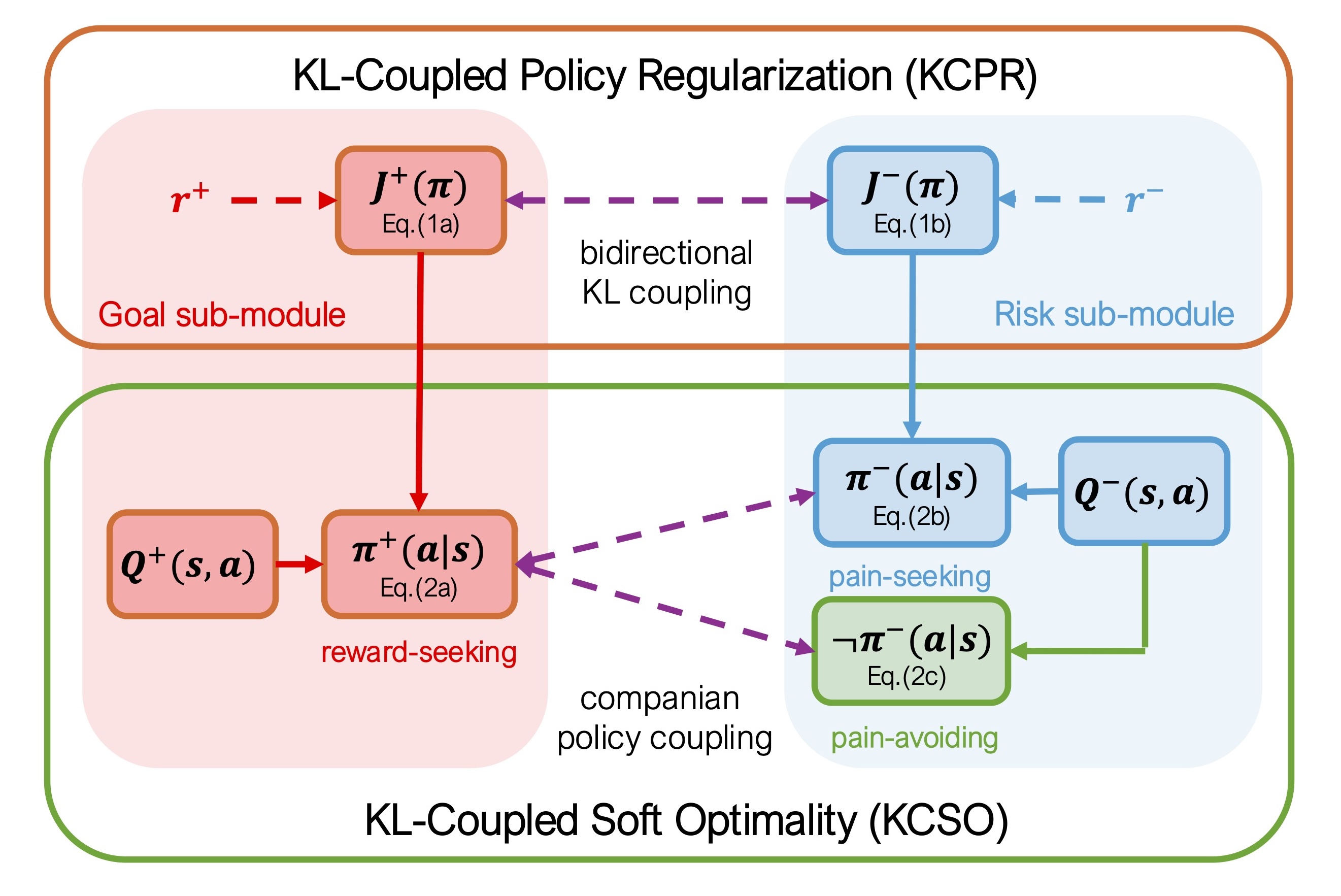}
\caption{Overview of the proposed framework. KCPR introduces bidirectional KL coupling at the objective level, where the reward-seeking and risk-related subsystems mutually regularize one another through coupled optimization objectives. Based on KCPR, KCSO derives a pair of companion policies with dynamically learned behavioral priors, establishing policy-level coordination between the two motivational subsystems.}
\label{arch}
\end{figure}

Experiments in both grid-world and deep robotic navigation tasks demonstrate that KCPR and the resulting KCSO framework enable structured interactions between reward-seeking and punishment-related learning. These interactions lead to safer and more risk-aware behaviors, more stable value propagation and learning dynamics, and competitive navigation performance compared with DQN, SQL and softDMP. Together, the results highlight the importance of policy-level coordination as an effective mechanism for balancing efficiency and safety in decomposed reinforcement learning.



More broadly, this work highlights a fundamental question in modern AI and robotics: how multiple internal objectives should be coordinated within a single learning system. 
While current foundation-model and vision-language-action approaches 
\citep{bommasani2021opportunities, brohan2022rt, zitkovich2023rt, black2024pi_0}
primarily focus on integrating perception, language, and action, comparatively less attention has been paid to the organization of internal motivational processes.
More generally, the proposed KCPR framework may provide a foundation for studying interactions among additional motivational signals, such as safety, curiosity, or uncertainty, in future embodied agents.

Our main contributions are:
\begin{enumerate}

\item \textbf{KL-Coupled Policy Regularization (KCPR).}
We introduce a new regularization principle in which companion policies act as dynamically learned priors, enabling policy-level interactions between reward and punishment pathways.

\item \textbf{KL-Coupled Soft Optimality (KCSO) and klDMP.}
We derive KCSO as the soft-optimality formulation of KCPR and develop klDMP as its practical realization within the RPRL framework.

\item \textbf{Stabilization mechanisms.}
We introduce a softening criterion for stable KL-coupled learning and provide an empirical analysis of replay-buffer designs for balancing reward- and punishment-driven updates.

\item \textbf{Deep robotic realization}
Through grid-world and robotic navigation experiments, we demonstrate that KL coupling improves safety, convergence stability, and risk-aware behavior compared with DQN, SQL and softDMP.

\end{enumerate}

\section{Related Work}
\label{sec:related}

\subsection{Reward-Punishment RL}
\label{sec:rprl}

Early modular reinforcement learning investigated learning separate value functions for different reward components or sub-goals \citep{Karlsson1997a, Humphrys1996a, Sprague2003a}. More recent methods, including Hybrid Reward Architecture (HRA) \citep{hra} and Reward Decomposition with Representation Disentanglement (RD$^2$) \citep{Lin2020c}, extended this idea to deep reinforcement learning by decomposing complex rewards into additive components. 

Reward–Punishment Reinforcement Learning (RPRL) instead separates environmental feedback into positive and negative channels, inspired by the dual valuation systems observed in biological decision making. Early formulations \citep{okada2001two, lowe2013exploring} employed separate actor–critic architectures for positive and negative rewards, while more recent methods, such as Reward–Punishment Actor–Critic \citep{rlac}, demonstrated their effectiveness in robotic manipulation. RPRL has also been used as a computational model of neurological and psychiatric disorders \citep{split, liebenow2022computational}.

Among RPRL methods, MaxPain \citep{Elfwing2017a} and Deep MaxPain (DMP) \citep{Wang2017b} most effectively leverage the punishment pathway. 
MaxPain introduces a ``min'' Bellman operator for negative rewards, enabling sharper propagation of aversive signals compared to standard value decomposition methods such as HRA. DMP further shows that directly optimizing negative return improves the shaping of positive value learning through shared dynamics. SoftDMP \citep{Wang2024} further introduced entropy regularization, replacing the hard max/min operators with mellow-max/mellow-min. This soft treatment reduces greedy exploitation of action values, and therefore improves robustness and sample efficiency. 

Despite these advances, existing RPRL methods optimize reward and punishment pathways largely independently, with interactions limited to value aggregation or entropy regularization. Consequently, policy-level coordination between motivational systems remains largely unexplored.

\subsection{Regularized RL}
\label{sec:rrl}

Kullback–Leibler (KL) divergence \citep{kl} plays a central role in modern reinforcement learning and is commonly used either to derive soft-optimal policies or to constrain policy updates during optimization.

A major line of work employs KL regularization to soften value updates and encourage stochastic exploration. Dynamic Policy Programming (DPP) \citep{Azar2012a} introduces a KL penalty relative to a baseline policy, yielding a log-sum-exp Bellman operator that stabilizes temporal-difference learning. G-learning \citep{Fox2016a} further replaces the baseline with a uniform prior, leading to a connection with the mellow-max operator \citep{Asadi2017a}. This perspective underlies Soft Q-Learning (SQL) \citep{Haarnoja2017a} and Soft Actor–Critic (SAC) \citep{Haarnoja2018a}, where KL regularization to a uniform prior is equivalent to entropy maximization. SoftDMP \citep{Wang2024} extends these ideas to RPRL.

A second line of work uses KL divergence as a trust-region constraint. Methods such as REPS \citep{peters2010relative}, TRPO \citep{schulman2015trust}, and PPO \citep{schulman2017proximal} constrain the divergence between successive policies to ensure stable policy improvement.

Among existing KL-regularized RL methods, our work is most closely related to soft-optimality approaches, but differs from existing approaches in two key aspects. First, KCPR replaces fixed reference distributions with dynamically learned companion policies. Second, KL regularization is used to coordinate interacting motivational subsystems rather than to encourage stochasticity, stabilize optimization, or constrain successive policy updates. These changes transform KL divergence from a policy regularizer into a mechanism for policy coordination.

\subsection{Policy Distillation and Multi-Task Regularization}
\label{sec:trl}

KL regularization is also central in multitask, transfer, and distillation-based reinforcement learning, where the goal is to share knowledge across tasks or agents. In Policy Distillation \citep{rusu2015policy}, a student policy is trained to imitate one or more teacher policies while optimizing its own objective, effectively compressing multiple specialized behaviors into a single network. Actor-Mimic \citep{parisotto2015actor} extends this idea by combining imitation with feature regression to improve cross-task generalization. Distral \citep{teh2017distral} further introduces a shared distilled policy that captures common behavioral structure across tasks while regularizing task-specific policies toward this shared prior.

In these methods, KL regularization facilitates knowledge sharing across different tasks or agents. By contrast, KCPR applies KL regularization between companion policies within the same MDP, where the policies optimize complementary reward components and co-adapt throughout learning. Consequently, KL divergence serves as a mechanism for coordinating interacting motivational systems rather than transferring knowledge across tasks.

\subsection{Relationship to Classical RL}
\label{sec:crl}

Our method is conceptually related to Expected SARSA \citep{van2009theoretical}, where value updates take an expectation over a policy rather than a greedy maximization. In klDMP, this expectation is instead defined by a learned companion policy. Unlike Expected SARSA, which evaluates the same policy used for sampling, klDMP remains off-policy because the expectation distribution is provided by a separately learned companion policy rather than the behavior policy.

The deep implementation follows a standard actor–critic-style architecture with a shared encoder and separate value and policy heads, similar to A3C \citep{mnih2016asynchronous}, SAC \citep{Haarnoja2018a}, and PPO \citep{schulman2017proximal}. Unlike conventional actor–critic methods, however, the policy heads are not optimized by policy gradients. Instead, they are directly supervised to match the analytically derived KCSO policies, yielding an explicit coupling between value estimation and policy formation.

\subsection{Safe RL}
\label{sec:safeRL}

Safe Reinforcement Learning aims to maximize cumulative rewards while explicitly enforcing safety constraints throughout learning \citep{Garcia2015a, gu2022review}. It is commonly formulated as a Constrained Markov Decision Process (CMDP), where safety is represented as an explicit cost function or risk constraint. In contrast, RPRL treats safety implicitly through negative rewards, allowing failures during exploration to improve subsequent value learning.

These paradigms therefore differ in their treatment of safety. Safe RL emphasizes preventing constraint violations and often provides theoretical or probabilistic safety guarantees, whereas RPRL relies on trial-and-error learning in which failures serve as informative training signals. Accordingly, Safe RL focuses on preventing failures, whereas RPRL focuses on learning from them.

\section{KL-regularized Deep MaxPain}
\label{sec:method}

\subsection{KL-Coupled Policy Optimization}
\label{sec:kcpo}

We consider an infinite-horizon discrete-time Markov Decision Process (MDP) defined by the tuple $(\mathcal{S, A, P, R},\gamma)$, where $\mathcal{S}$ denotes the state space, $\mathcal{A}$ the discrete action space, $\mathcal{P}$ the transition probability, $\mathcal{R}$ the reward function, and $\gamma \in [0,1)$ the discount factor. At each time step $t$, the agent selects an action $a_t \in \mathcal{A}$ in response to the current state $s_t \in \mathcal{S}$, according to a stochastic behavior policy $\pi(a_t \mid s_t)$. The environment then transitions to the next state $s_{t+1} \sim \mathcal{P}(\cdot \mid s_t,a_t)$, yielding a scalar reward from the reward function $r_t = \mathcal{R}(s_t, a_t, s_{t+1})$ as an evaluation signal.

Figure~\ref{arch} provides an overview of the proposed framework. At the objective level, KL-Coupled Policy Regularization (KCPR) introduces bidirectional coupling between the reward-seeking and punishment-related optimization objectives. Based on KCPR, KL-Coupled Soft Optimality (KCSO) derives the corresponding companion policies, enabling policy-level interactions between the two motivational subsystems. 
The remainder of this section formalizes these two components. Following the RPRL paradigm, the environmental reward is decomposed into positive and negative components,
\begin{align*}
  r^+ = \max (\mathcal{R}, 0), \quad r^- = \min (\mathcal{R}, 0),
\end{align*}
where $r^+ \ge 0$ represents the goal-seeking reward signal and $r^- \le 0$ represents the punishment signal.

At the objective level, KCPR replaces the fixed reference prior in conventional KL regularization with a dynamically learned companion policy. Formally, given a reward function $r$, a companion policy $\pi_{\text{c}}$, and a regularization coefficient $\eta$, KCPR defines the following objective:
\begin{align*}
  J(\pi;\pi_{\text{c}},r,\eta)=\mathbb{E}_{\pi,\mathcal{P}} \left[\sum_{t=0}^{\infty} \gamma^t \left(r_t-\frac{1}{\eta} D_{\text{KL}} (\pi(\cdot|s_t)||\pi_{\text{c}}(\cdot|s_t)) \right) \right].
\end{align*}
For comparison, entropy-regularized methods such as softDMP use a fixed uniform prior,
\begin{align*}
  \pi_{\text{c}}=\pi_{\text{uniform}}.
\end{align*}
Within the reward–punishment setting considered in this work, the positive reward signal $r^+$ is regularized by the pain-avoiding policy $\neg\pi^-$, while the negative reward signal $r^-$ is regularized by the goal-seeking policy $\pi^+$:
\begin{subequations} 
  \label{eq:kcpr_objectives} 
  \begin{align} 
    J^+(\pi) &= J(\pi;\neg\pi^-,r^+,\eta^+), \label{eq:kcpr_objectives_a}\\ 
    J^-(\pi) &= J(\pi;\pi^+,r^-,\eta^-). \label{eq:kcpr_objectives_b} \end{align} 
\end{subequations}
The corresponding optimal policies are defined as
\begin{align*}
  \pi^+=\arg \max_{\pi} J^+(\pi), \quad \pi^-=\arg \min_{\pi} J^-(\pi).
\end{align*} 
The asymmetric max–min formulation plays a key role in KCPR. Maximizing $J^+$ encourages the goal-seeking policy to align with the pain-avoiding companion prior, while minimizing $J^-$ yields a pain-seeking policy whose behavior remains differentiated from the goal-seeking policy. 

The coupled objectives above define the optimization problem. We next derive their corresponding soft-optimal solutions, referred to as KL-Coupled Soft Optimality (KCSO). Formally, consider a policy regularized by a companion policy $\pi_c$. The KCPR objective admits the following soft-optimal solution:
\begin{align*}
  \pi^*(a|s) = \frac{\pi_{\mathrm{c}}(a|s) \exp\big[\eta Q(s,a)\big]}{\sum_{a'} \pi_{\mathrm{c}}(a|s) \exp\big[\eta Q(s,a')\big]}.
\end{align*}
KCSO treats the prior as a dynamically learned companion policy:
\begin{subequations} 
  \label{eq:optimal_policy} 
  \begin{align} 
    \pi^+ (a \mid s) &= \frac{\neg \pi^- (a \mid s)\exp[\eta^+Q^+(s,a)]} {\sum_{a'}\neg \pi^-(a' \mid s)\exp[\eta^+Q^+(s,a')]}, \label{eq:optimal_policy_a} \\ 
    \pi^- (a \mid s) &= \frac{\pi^+(a \mid s)\exp[\eta^-Q^-(s,a)]} {\sum_{a'}\pi^+(a' \mid s)\exp[\eta^-Q^-(s,a')]}, \label{eq:optimal_policy_b} \\ 
    \neg\pi^- (a \mid s) &\triangleq \frac{\pi^+(a \mid s)\exp[-\eta^-Q^-(s,a)]} {\sum_{a'}\pi^+(a' \mid s)\exp[-\eta^-Q^-(s,a')]}, \label{eq:optimal_policy_c} 
  \end{align} 
\end{subequations}
where Eq.~\eqref{eq:optimal_policy_a} defines the goal-seeking policy $\pi^+$ using the pain-avoiding policy $\neg \pi^-$ as its companion prior. 
Eq.~\eqref{eq:optimal_policy_b} defines the pain-seeking policy $\pi^-$ using the goal-seeking policy $\pi^+$ as its companion prior, while Eq.~\eqref{eq:optimal_policy_c} defines the complementary pain-avoiding policy $\neg \pi^-$ under the same companion prior $\pi^+$ by reversing the sign of the punishment-related action values.
The sign of $\eta$ determines whether the policy favors high- or low-value actions: $\eta^+>0$ yields the reward-seeking policy $\pi^+$, whereas $\eta^-<0$ yields the pain-seeking policy $\pi^-$. The complementary policy $\neg\pi^-$ favors actions associated with lower expected punishment and serves as the companion prior for the goal-seeking subsystem.

\subsection{Tabular Learning rules}
\label{sec:learning_rule}

The KCPR objective and the resulting KCSO principle naturally give rise to a practical reinforcement learning algorithm. In the tabular setting, we refer to this algorithm as klMP. The deep neural network realization introduced later is referred to as klDMP. This section first presents the tabular learning rules underlying klMP, which form the foundation of the deep implementation.

Under KCSO, value propagation is governed by a pair of KL-coupled Bellman operators associated with the reward-seeking and punishment-related subsystems. These operators are conditioned on the dynamically evolving companion policies derived in Eq.~\eqref{eq:optimal_policy}, allowing each subsystem to propagate value information under the behavioral guidance of its companion policy. For an experience tuple $(s,a,r,s')$, the KL-coupled Bellman targets are defined as:
\begin{subequations}
\label{eq:bellman_target}
\begin{align}
y^+=r^+ +  \frac{\gamma^+}{\eta^+}\log \sum_{a'} \neg \pi^-(a'|s)\exp [\eta^+ Q^+ (s',a')], \\
y^-=r^- + \frac{\gamma^-}{\eta^-} \log \sum_{a'} \pi^+(a'|s)\exp [\eta^- Q^- (s',a')].
\end{align}
\end{subequations}
where the first target propagates reward information under the pain-avoiding companion policy $\neg \pi^-$, while the second propagates punishment-related information under the goal-seeking companion policy $\pi^+$. 
The corresponding temporal-difference errors are
\begin{align*}
\delta^+=y^+-Q^+(s,a), \quad \delta^-=y^--Q^-(s,a).
\end{align*}
and the action-value functions are updated using standard temporal-difference learning,
\begin{subequations}
\label{eq:q_updating}
\begin{align}
Q^+(s,a) \leftarrow Q^+(s,a)+ \alpha^+ \delta^+, \\
Q^-(s,a) \leftarrow Q^-(s,a)+ \alpha^- \delta^-. 
\end{align}
\end{subequations}
where $\alpha^+$ and $\alpha^-$ denote the learning rates for the goal-seeking and punishment-related modules, respectively.

After each value update, the companion policies are renewed according to Eq.~\eqref{eq:optimal_policy}. Consequently, value estimation and policy formation are coupled throughout learning: updated value functions determine new companion policies, which in turn define subsequent KL-coupled Bellman targets.

\subsection{Behavior policy}
\label{sec:behavior_policy}

The KCSO operator defines the soft-optimal policies associated with the goal-seeking and punishment-related subsystems. However, directly executing these policies may lead to insufficient exploration, particularly during the early stages of learning when both value estimates and companion priors remain inaccurate. Therefore, klDMP constructs a separate behavior policy for data collection while preserving the coupling structure induced by KCPR and KCSO.
To regulate exploration during learning, an annealable temperature parameter $\tau>0$ is introduced. The resulting softened policies are defined as
\begin{subequations}
\label{eq:sub_policies}
\begin{align}
  \tilde{\pi}^+ (a \mid s) \propto \neg \pi^- (a \mid s)^{1/\tau} \cdot \exp [\eta Q^+ (s, a)/\tau],  \\
  \neg \tilde{\pi}^- (a \mid s) \propto \pi^+ (a \mid s)^{1/\tau} \cdot \exp [ -\eta Q^-(s, a)/\tau], \\
  \tilde{\pi}^- (a \mid s) \propto \pi^+ (a \mid s)^{1/\tau} \cdot \exp [ \eta Q^-(s, a)/\tau].
\end{align}
\end{subequations}
As $\tau \to 1$, the softened policies converge to their original soft-optimal forms: $\tilde{\pi}^+ \to \pi^+$, $\tilde{\pi}^- \to \pi^-$, and $\neg\tilde{\pi}^- \to \neg\pi^-$ (Eq.\eqref{eq:optimal_policy}). Conversely, as $\tau \to \infty$, all sub-policies approach uniform distributions, ensuring adequate exploration during early training. The overall behavior policy is defined as a mixture of the softened sub-policies: 
\begin{equation}
\bar{\pi}(a \mid s) = w \tilde{\pi}^+ (a \mid s)
  + (1 - w) \neg \tilde{\pi}^- (a \mid s),
  \label{eq:policy_mixture}
\end{equation}
where the mixing weight $w \in [0,1]$ controls the relative influence of the goal-seeking and pain-avoiding sub-modules. Similar to DMP, the mixing weight $w$ may either be fixed or adapted online \citep{Wang2021a} according to the relative strengths of the two motivational systems.

Note that the behavior policy is used solely for data collection, whereas policy optimization remains governed by KCSO. This separation allows exploration to be adjusted independently without altering the KL-coupled optimality principle.


\subsection{Softening companion priors}
\label{sec:soften_prior}

The effectiveness of KCPR relies on the use of learned companion policies as dynamic priors. However, during the early stages of learning, these companion policies are derived from immature value estimates and may therefore become prematurely concentrated on suboptimal actions. When such overconfident priors are incorporated into the KCSO operator and the corresponding Bellman backups, value propagation through alternative actions can be significantly suppressed, leading to unstable learning dynamics. To alleviate this issue, we introduce a softening mechanism that interpolates the companion priors with a uniform distribution:
\begin{subequations}
  \label{eq:soften}
  \begin{align}
    \neg \pi^-_{\text{soften}}(a \mid s)=\epsilon \pi_{\text{uniform}}+(1-\epsilon) \neg \pi^-(a \mid s), \\
\pi^+_{\text{soften}}(a \mid s)=\epsilon \pi_{\text{uniform}}+(1-\epsilon) \pi^+(a \mid s).
  \end{align}
  \end{subequations}
where $\pi_{\text{uniform}}=\frac{1}{|\mathcal{A}|}$, and $\epsilon \in [0,1]$ controls the degree of softening. 
The proposed formulation establishes a continuous transition between entropy-based and KL-coupled regularization. When $\epsilon=1$, the companion priors reduce to a uniform distribution, recovering entropy regularization. When $\epsilon=0$, the original learned companion priors are recovered, yielding the standard klDMP formulation. Note that the softened companion priors are used consistently throughout learning, including the KL-regularized Bellman backups of $Q^+$ and $Q^-$, the computation of KCSO policies, and the construction of the behavior policy. 

\subsection{Separate replay buffer}
\label{sec:replay_buffer}

To alleviate experience imbalance between the two motivational subsystems, we adopt the separate replay buffer scheme introduced in softDMP \citep{Wang2024}. In practice, the pain-avoiding component often dominates data collection, resulting in insufficient training samples for the pain-seeking value function $Q^-$. To ensure balanced learning, experiences are assigned to separate replay buffers using a discriminator: 
\begin{equation}
D(s,a,r,s')=\frac{\tilde{\pi}^- (a \mid s)}{\tilde{\pi}^- (a \mid s)+\tilde{\pi}^+ (a \mid s)}
\label{eq:discriminator}
\end{equation}
which quantifies the relative contribution of the pain-seeking policy to a sampled transition. Transitions with higher discriminator values are stored in the negative replay buffer $\mathcal{D}^-$ for updating $Q^-$, while the remaining samples are assigned to the positive replay buffer $\mathcal{D}^+$ for updating $Q^+$.

By decoupling experience storage, each subsystem receives training data that better reflects its behavioral objective. This improves the propagation of punishment-related value information, facilitates more balanced learning between the two motivational subsystems, and reduces interference between reward-seeking and punishment-related updates.

\subsection{Deep neural network realization}
\label{sec:deep_realization}

\begin{figure}[H]
  \centering
  \includegraphics[width=0.8\hsize]{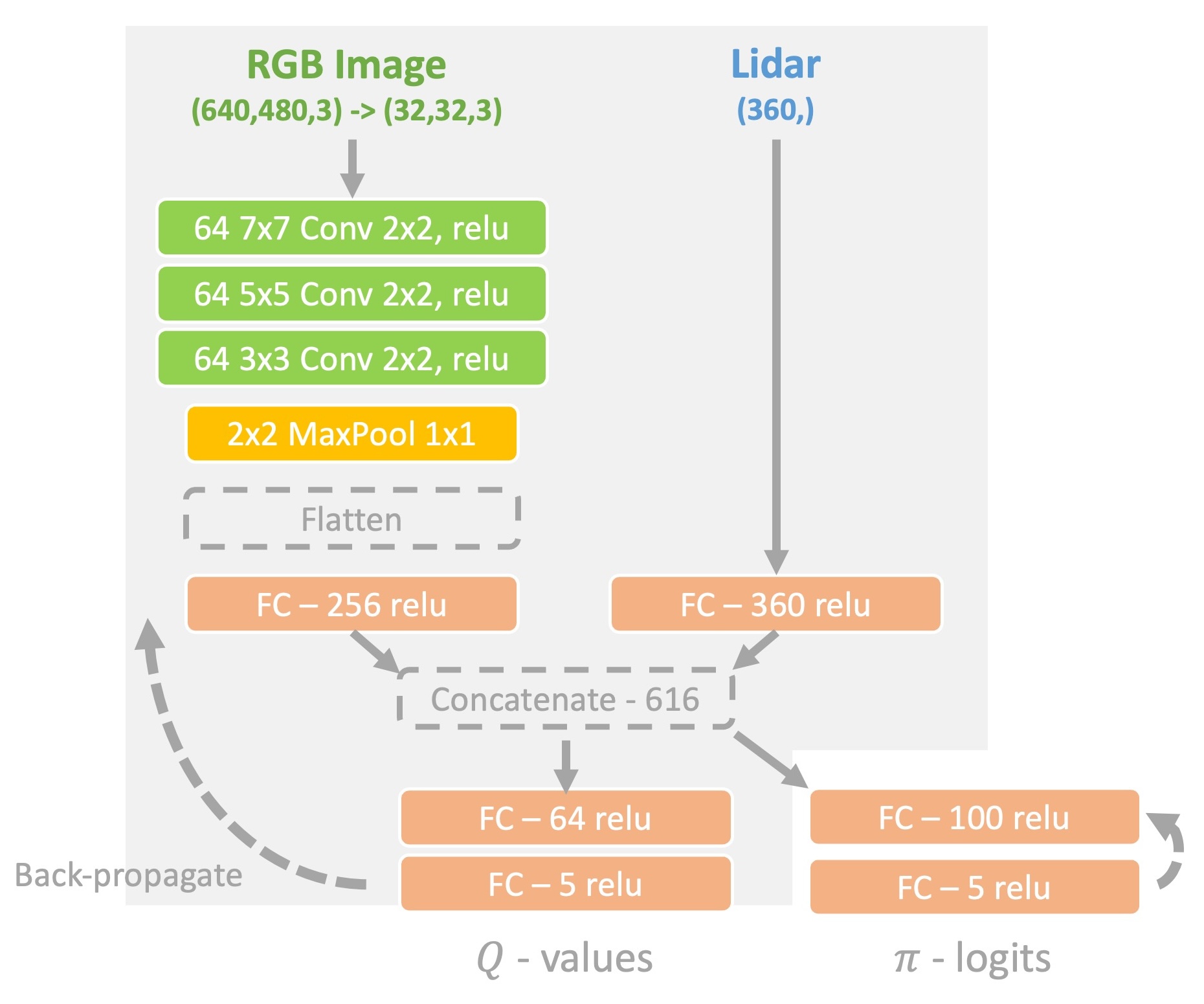}
  \caption{Dual-network realization of the proposed klDMP framework.}
  \label{networks}
\end{figure}

In the deep implementation (klDMP), each motivational subsystem is realized by a neural network with separate value and policy heads (Figure~\ref{networks}). Both modules share the same multimodal perception pipeline, where RGB images and LiDAR measurements are encoded into a shared feature representation before branching into value and policy heads. The goal-seeking network outputs $(Q^+,\pi^+)$, while the punishment-related network outputs $(Q^-,\neg\pi^-)$. The value heads propagate gradients through the entire network, ensuring that representation learning is driven by temporal-difference errors. In contrast, the policy heads are restricted to backpropagation only up to the shared feature layer, decoupling policy optimization from low-level feature learning stability.
\begin{algorithm}[!t]
\caption{klDMP Learning Algorithm}
\label{alg:kldmp}
\begin{algorithmic}[1]
\REQUIRE hyperparameters $\alpha^\pm,\gamma^\pm,\eta^\pm,\epsilon$
\STATE Initialize online value and policy networks $Q^+_{\theta^+}, Q^-_{\theta^-},\pi^+_{\phi^+}, \neg\pi^-_{\phi^-}$ with parameter $\theta^+, \theta^-, \phi^+, \phi^-$
\STATE Initialize target networks $\bar{\theta}^+ \leftarrow \theta^+$, $\bar{\theta}^- \leftarrow \theta^-$, $\bar{\phi}^+ \leftarrow \phi^+$, $\bar{\phi}^- \leftarrow \phi^-$
\STATE Initialize replay buffers $\mathcal{D}^+$ and $\mathcal{D}^-$

\FOR{each episode}
    \STATE Initialize state $s$
    \WHILE{$s$ is not terminal}
        \STATE Construct the behavior policy $\bar{\pi}(a|s)$ using Eq.~\eqref{eq:soften} and Eq.~\eqref{eq:policy_mixture}
        \STATE Sample action $a \sim \bar{\pi}(\cdot|s)$
        \STATE Execute action $a$, observe $r^+, r^-, s',d$
        \STATE Assign $(s,a,r^+,s',d)$ to $\mathcal{D}^+$ and $(s,a,r^-,s',d)$ to $\mathcal{D}^-$ using Eq.~\eqref{eq:discriminator}
        \STATE Sample mini-batches from $\mathcal{D}^+$ and $\mathcal{D}^-$
        \STATE Compute $y_{Q^+}, y_{Q^-}$ using Eq.~\eqref{eq:deep_q_target}
        \STATE Update $\theta^+, \theta^-$ by minimizing $L_{Q^+}, L_{Q^-}$ using Eq.~\eqref{eq:loss_q}
        \STATE Compute $y_{\pi^+}, y_{\neg\pi^-}$ using Eq.~\eqref{eq:deep_pi_target}
        \STATE Update $\phi^+, \phi^-$ by minimizing $L_{\pi^+}, L_{\neg\pi^-}$ using Eq.~\eqref{eq:loss_pi}
        
        \IF{target update condition is satisfied}
            \STATE Update target networks:
            $\bar{\theta}^+ \leftarrow \theta^+$,
            $\bar{\theta}^- \leftarrow \theta^-$,
            $\bar{\phi}^+ \leftarrow \phi^+$,
            $\bar{\phi}^- \leftarrow \phi^-$
        \ENDIF
        
        \STATE $s \leftarrow s'$
    \ENDWHILE
\ENDFOR
\end{algorithmic}
\end{algorithm}

Let $Q^+_{\theta^+}$ and $Q^-_{\theta^-}$ denote the value function estimators parameterized by $\theta^+$ and $\theta^-$, and let $\pi^+_{{\phi}^+}$ and $\neg\pi^-_{{\phi}^-}$ denote the corresponding policy heads parameterized by $\phi^+$ and $\phi^-$. The value losses are defined as:
\begin{subequations}
  \label{eq:loss_q}
\begin{align}
  L_{Q^+} = \mathbb{E}_{\mathcal{D}^+} \left[ \left( Q^+_{\theta^+} (s, a) - y_{Q^+} \right)^2 \right], \\
  L_{Q^-} = \mathbb{E}_{\mathcal{D}^-} \left[ \left( Q^-_{\theta^-} (s, a) - y_{Q^-} \right)^2 \right].
\end{align}
\end{subequations}
The deep implementation follows the same KL-coupled learning rules as the tabular algorithm. Specifically, the Bellman backups introduced in Eq.~\eqref{eq:bellman_target} are applied to neural value functions and target networks to construct the training targets:
\begin{subequations}
\label{eq:deep_q_target}
\begin{align}
  y_{Q^+}  \triangleq r^+ + \frac{(1 - d) \gamma^+}{\eta^+} 
  \log \sum_{a'} \neg \pi^-_{\bar{\phi}^- \text{soften}} (a' \mid s') \exp \left[\eta^+ Q^+_{\bar{\theta}^+} (s',a') \right],\\
  y_{Q^-}  \triangleq r^- + \frac{(1 - d) \gamma^-}{\eta^-} 
  \log \sum_{a'}  \pi^+_{\bar{\phi}^+ \text{soften}} (a' \mid s') \exp \left[\eta^- Q^-_{\bar{\theta}^-} (s',a') \right].
\end{align}
\end{subequations}
Here, $\mathcal{D}^+$ and $\mathcal{D}^-$ denote the positive and negative replay buffers, respectively. 
Compared with the tabular formulation, the deep implementation additionally incorporates target networks and terminal-state masking through $(1-d)$, where $d$ indicates episode termination. The target networks $\bar{\theta}^+,\bar{\theta}^-,\bar{\phi}^+,\bar{\phi}^-$ are periodically updated copies of the online networks and are used to stabilize KL-coupled value propagation during training.
The policy heads are trained by matching analytically derived KL-optimal targets:
\begin{subequations}
  \label{eq:loss_pi}
  \begin{align}
  L_{\pi^+} = \mathbb{E}_{\mathcal{D}^+}\left[D_{\text{KL}}(y_{\pi^+}(\cdot \mid s) || \pi^+_{\phi^+}(\cdot\mid s)) \right], \\
  L_{\neg\pi^-} = \mathbb{E}_{\mathcal{D}^-}\left[D_{\text{KL}}(y_{\neg\pi^-}(\cdot \mid s) || \neg\pi^-_{\phi^-}(\cdot\mid s)) \right].
\end{align}
\end{subequations}
The corresponding target policies are defined as:
\begin{subequations}
\label{eq:deep_pi_target}
\begin{align}
  y_{\pi^+}(a \mid s) = \frac{\neg \pi^-_{\bar{\phi}^-\text{soften}}(a \mid s) \exp [\eta^+ Q^+_{\bar{\theta}^+}(s,a)]}{\sum_{a'} \neg \pi^-_{\bar{\phi}^-\text{soften}}(a \mid s) \exp [\eta^+ Q^+_{\bar{\theta}^+}(s,a')]}, \\
  y_{\neg\pi^-}(a \mid s) = \frac{\pi^+_{\bar{\phi}^+\text{soften}}(a \mid s) \exp [-\eta^- Q^-_{\bar{\theta}^-}(s,a)]}{\sum_{a'} \pi^+_{\bar{\phi}^+\text{soften}}(a \mid s) \exp [-\eta^- Q^-_{\bar{\theta}^-}(s,a')]}. 
\end{align}
\end{subequations}
Figure~\ref{networks} and Algorithm~\ref{alg:kldmp} summarize the complete deep realization of the proposed framework. 

\section{Experiments}
\label{sec:experiments}

\subsection{Grid-world}

We evaluate klMP in low-dimensional grid-world environments to analyze how KCPR influences policy coupling, value propagation, and learning dynamics under reward–punishment decomposition. Specifically, we use a 9×9 U-maze \citep{Wang2021a,Wang2024} and a 36×19 Three-room maze to investigate three key components of the proposed framework:

\begin{itemize}
\item \textbf{Effects of KL regularization:}
how companion-policy regularization shapes policy coupling and the trade-off between goal-seeking and risk-avoiding behaviors.

\item \textbf{Effects of companion-prior softening:}
how mitigating overconfident companion priors stabilizes value propagation during early learning.

\item \textbf{Effects of separate replay buffers:}
how separating reward- and punishment-related experiences improves learning stability and punishment-related value propagation.
\end{itemize}

\subsubsection{Effects of regularization}

To isolate the effect of KL-regularization, we first construct a baseline setting consisting of two independently trained agents under a fixed environment. The goal-seeking agent receives a reward of $+1$ upon reaching the goal, while the pain-avo
iding agent receives a penalty of $-0.1$ when colliding with obstacles. All other transitions yield zero reward, and episodes terminate upon reaching the goal. Both agents are trained using model-based value iteration (QVI), following a MaxPain-style update scheme where goal-directed learning employs a max operator and pain-avoiding learning employs a min operator. The resulting optimal policies are obtained via a low-temperature softmax over action values.

\begin{figure}[H]
\centering
\includegraphics[width=0.6\hsize]{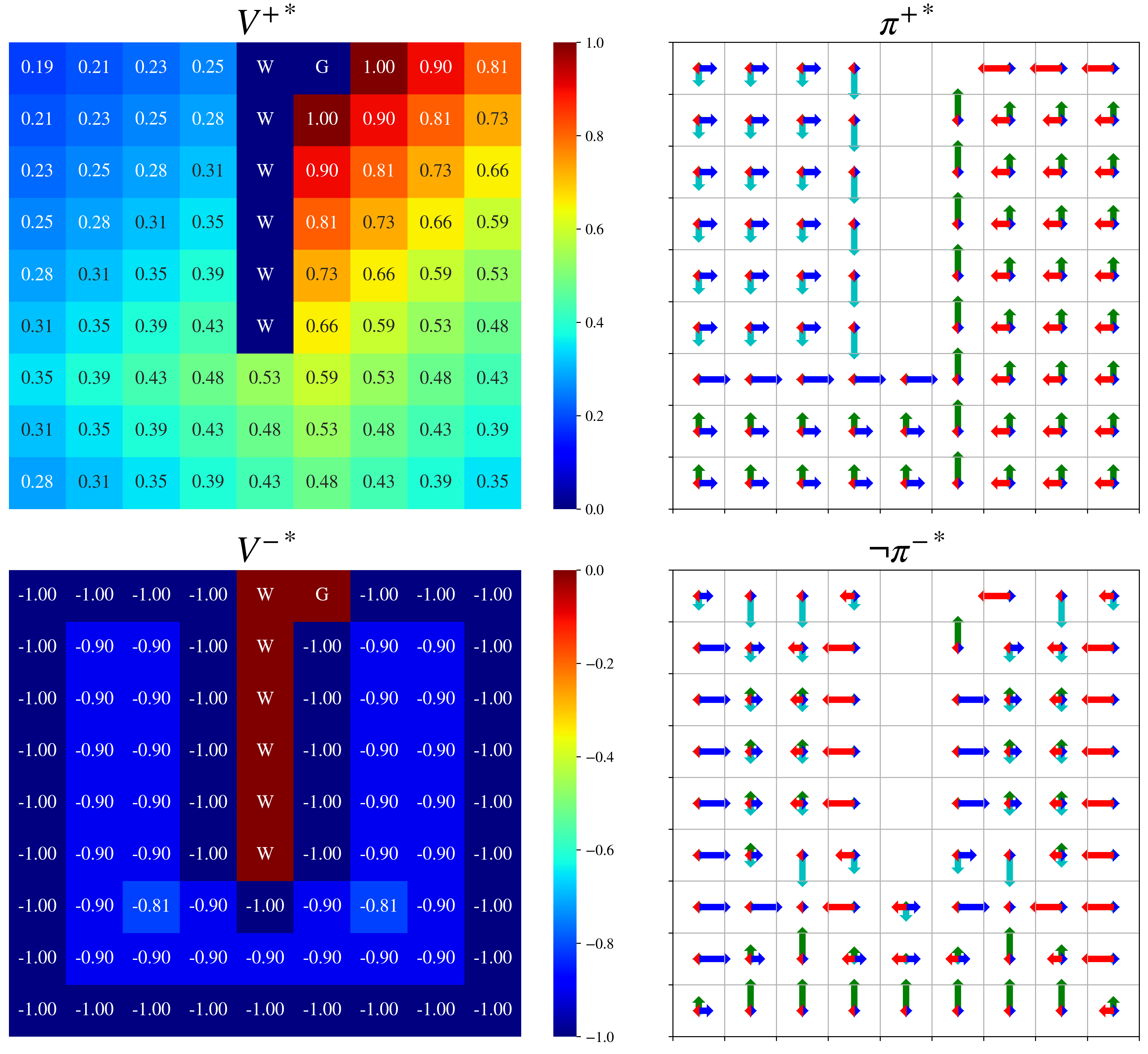}
\caption{Optimal state-value functions and the corresponding reward-seeking and pain-avoiding policies learned independently. These policies are subsequently used as fixed companion priors in the klQVI experiments to analyze the effects of KL-Coupled Policy Regularization.}
\label{greedy_q_pi}
\end{figure}

Figure \ref{greedy_q_pi} shows the resulting value functions and policies. The goal-seeking agent learns a monotonic value landscape toward the goal, resulting in shortcut-oriented navigation behavior. In contrast, the pain-avoiding agent forms structured low-value regions around obstacles, effectively inducing safety zones that shape cautious navigation behavior. To study cross-subsystem interaction, we next introduce KL-coupling by treating the independently learned policies $\pi^{+*}$ and $\neg\pi^{-*}$ as fixed companion priors in a second round of QVI, referred to as klQVI. In this setting, each Bellman update is regularized toward the behavior of the opposite subsystem, as defined in Eq.~\eqref{eq:bellman_target}.

\begin{figure}[H]
\centering
\begin{subfigure}{0.9\linewidth}
    \centering
    \includegraphics[width=0.8\linewidth]{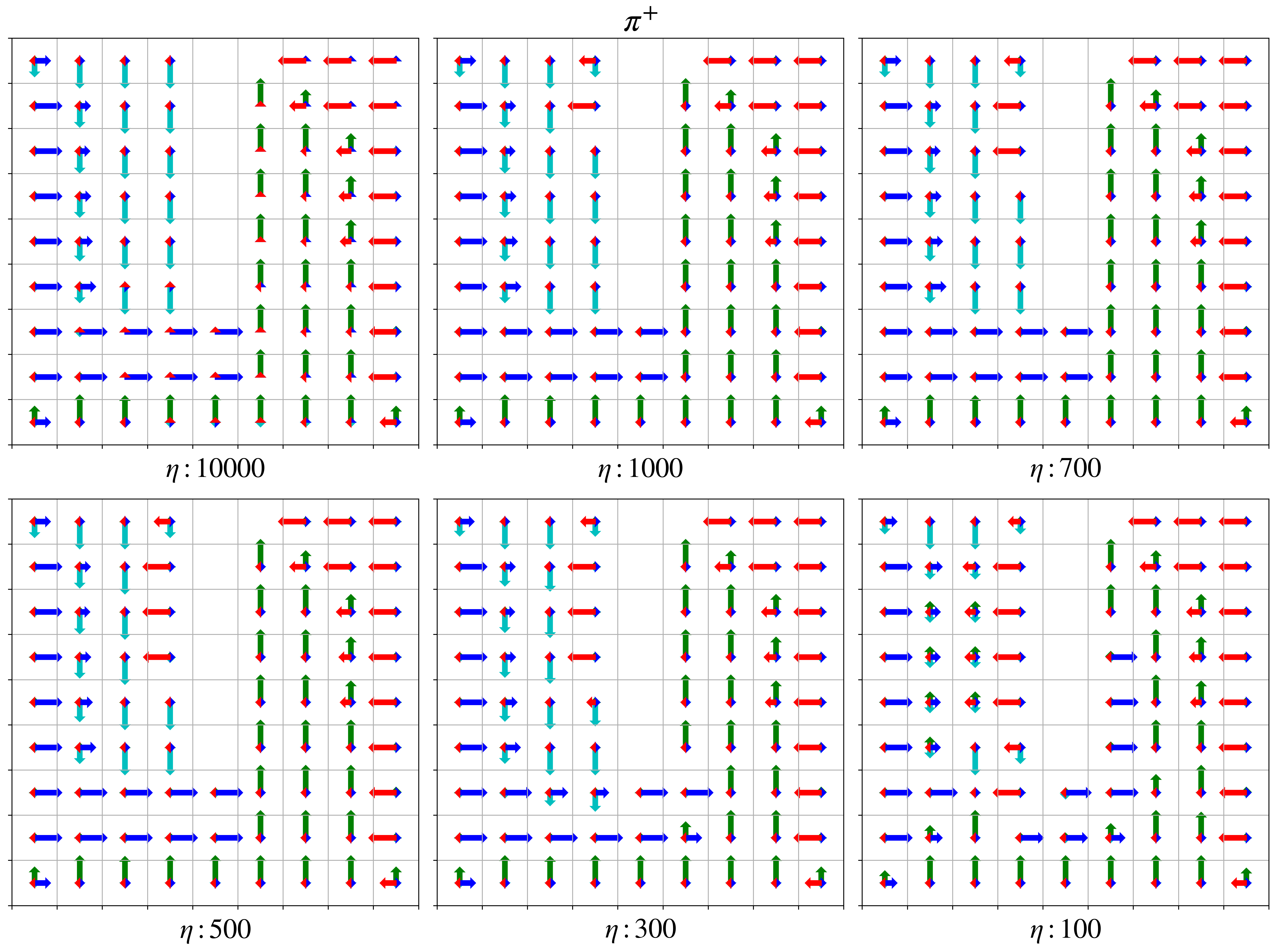}
    \caption{the optimal goal-seeking policy derived from $Q^+$ learned from klQVI with different $\eta^+$}
    \label{fig:subfig_a}
\end{subfigure}

\begin{subfigure}{0.9\linewidth}
    \centering
    \includegraphics[width=0.8\linewidth]{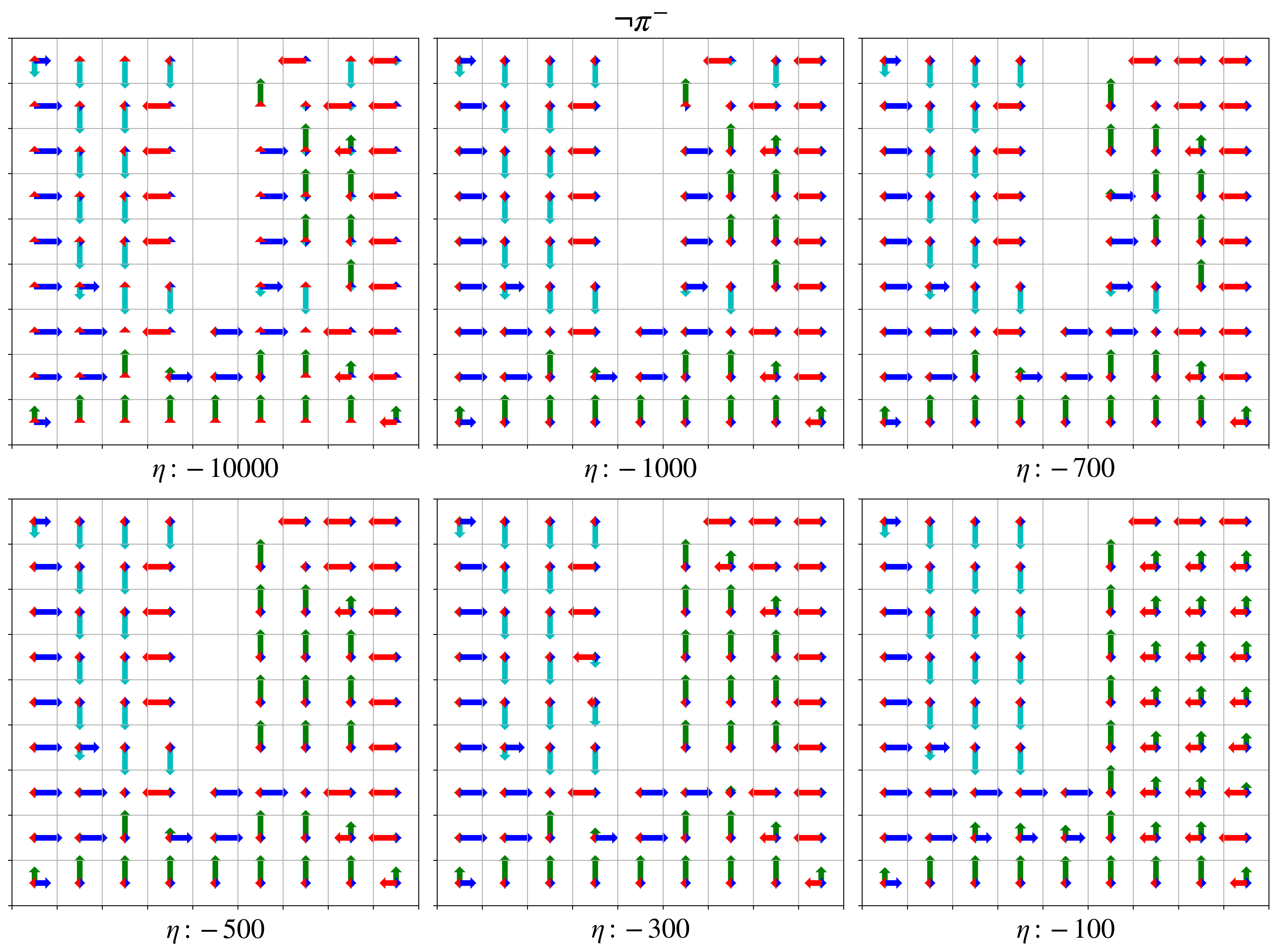}
    \caption{the optimal pain-avoiding policy derived from $Q^-$ learned from klQVI with different $\eta^-$}
    \label{fig:subfig_b}
\end{subfigure}

\caption{Optimal policies obtained from klQVI under different KL regularization strengths. It illustrates how KCPR controls policy coupling through the companion prior. Smaller values of $|\eta|$ increase the influence of the companion policy, leading to stronger behavioral blending between reward-seeking and pain-avoiding policies, whereas larger values recover behavior closer to independent learning.}
\label{kl_qvi_pi}
\end{figure}

Figure~\ref{kl_qvi_pi} illustrates the resulting optimal policies under different KL regularization strengths. A clear and systematic blending between goal-directed and risk-averse behaviors emerges as a function of $\eta$. When $\eta^+$ is large, the KL term becomes weak and the goal-seeking policy closely follows the standard QVI solution. As $\eta^+$ decreases, the influence of the pain-avoiding prior $\neg \pi^{-*}$ becomes stronger, leading to increasingly conservative navigation behavior. A symmetric trend is observed in the punishment-learning case, where smaller $|\eta^-|$ amplifies the influence of the goal-seeking prior.

These results provide empirical evidence for KCPR. By treating companion policies as learned priors, KCPR enables continuous behavioral shaping between reward-seeking and punishment-related subsystems. The resulting KCSO policies exhibit smooth transitions between independent and strongly coupled behaviors, demonstrating how policy-level coupling can regulate the exploration–safety trade-off.

\subsubsection{Effects of softening criterion}

We evaluate klDMP with and without the proposed softening criterion in a grid-world environment where reward and punishment signals coexist. The agent receives a reward of $+1$ upon reaching the goal and a penalty of $-0.1$ for collisions, with all other rewards set to zero. The goal state is absorbing. The algorithm maintains two action-value functions $Q^+$ and $Q^-$, together with their corresponding KL-coupled policies, which are updated jointly according to Eq.~\eqref{eq:optimal_policy} and Eq.~\eqref{eq:q_updating}.

\begin{figure}[H]
\centering
\begin{subfigure}{1\linewidth}
    \centering
    \includegraphics[width=1\linewidth]{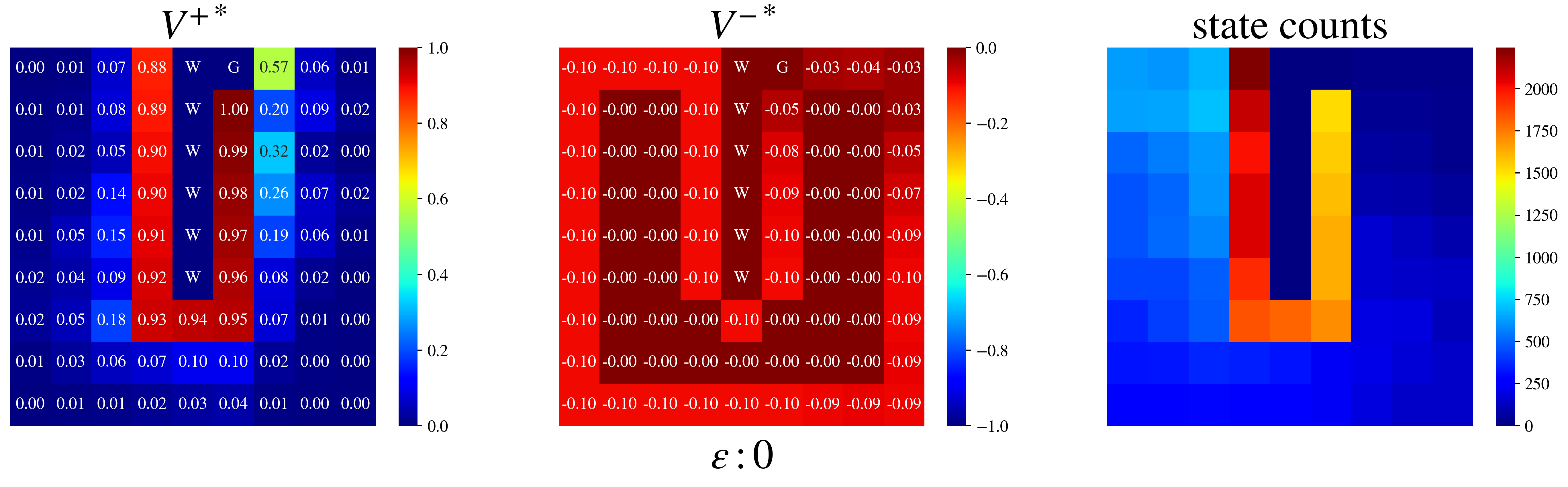}
    \caption{$\epsilon=0$ (klMP case)}
    \label{fig:kl_soften_a}
\end{subfigure}

\begin{subfigure}{1\linewidth}
    \centering
    \includegraphics[width=1\linewidth]{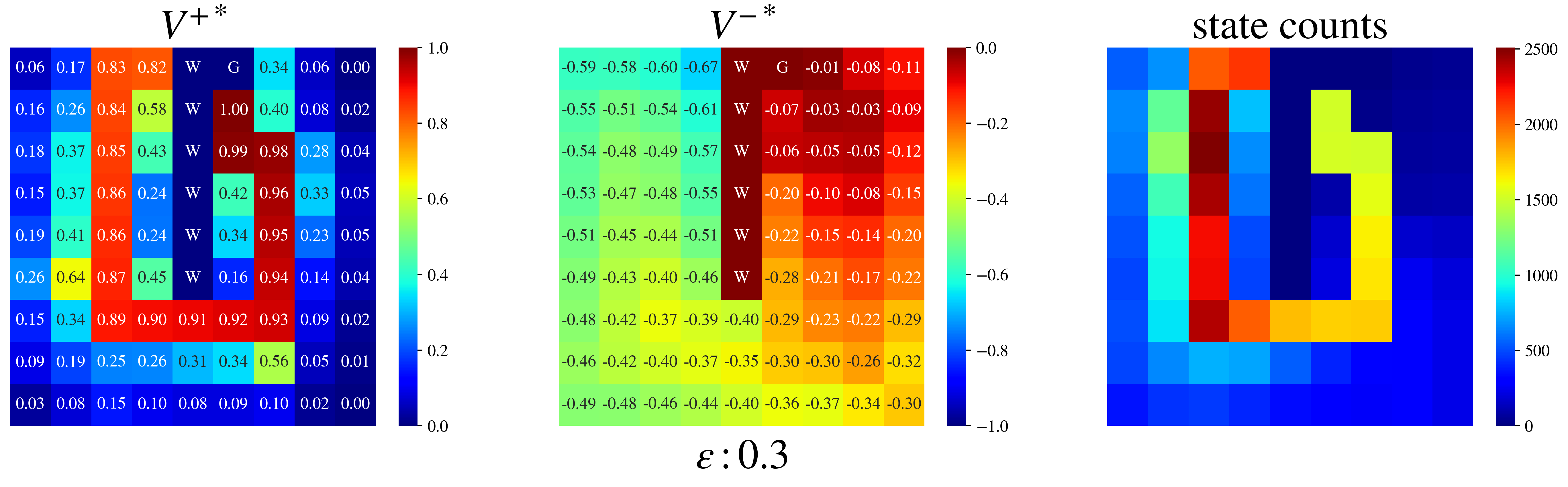}
    \caption{$\epsilon=0.3$ (softened klMP case)}
    \label{fig:kl_soften_b}
\end{subfigure}

\caption{Heatmaps of the learned state-value functions and state visitation counts with and without the proposed softening criterion. Without softening ($\epsilon=0$), the companion prior becomes overconfident during early learning, suppressing the propagation of punishment-related values and resulting in shortcut-oriented exploration. With softening ($\epsilon=0.3$), both reward and punishment values propagate more effectively, leading to a more balanced visitation distribution and safer navigation behavior.}
\label{kl_soften}
\end{figure}

Figure~\ref{kl_soften} compares pure klMP $(\epsilon=0)$ and softened klMP $(\epsilon=0.3)$ under identical KL regularization strength $(\eta^+,\eta^-)=(1000,-1000)$. Without softening, the rapidly sharpening companion priors lead to premature policy concentration, which in turn suppresses the propagation of negative value signals in $Q^-$. As a result, the agent exhibits shortcut-oriented behavior dominated by goal-seeking dynamics, with insufficient risk-awareness. In contrast, introducing softening preserves early-stage exploration by preventing overconfident companion policies. This allows more balanced propagation of both positive and negative value signals, resulting in a more structured representation of risk regions and improved trade-off between goal reaching and collision avoidance.

\begin{figure}[H]
\centering
\begin{subfigure}{0.48\linewidth}
    \centering
    \includegraphics[width=\linewidth]{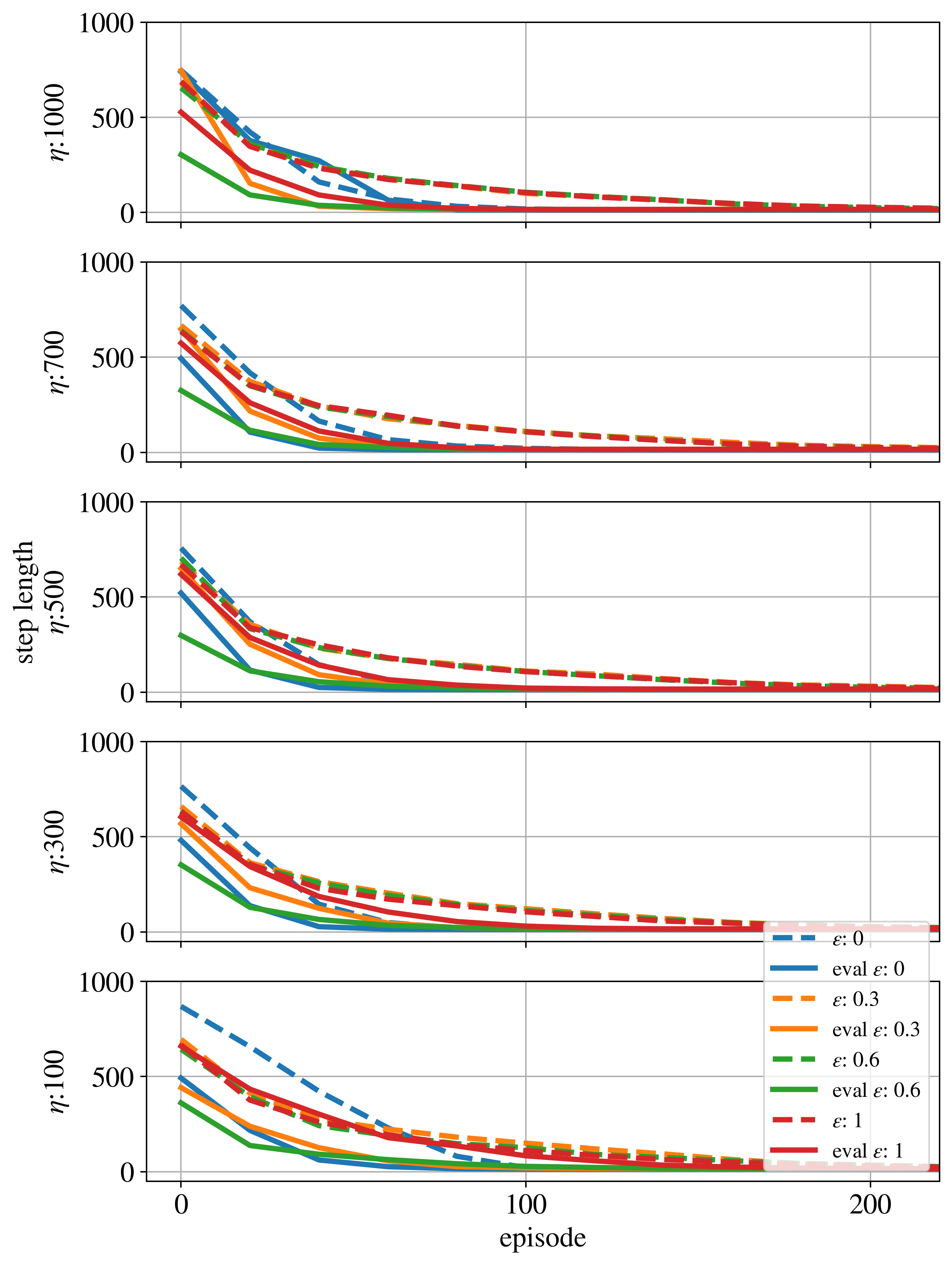}
    \caption{The step length convergence}
    \label{fig:kl_soften_c}
\end{subfigure}
\hfill
\begin{subfigure}{0.48\linewidth}
    \centering
    \includegraphics[width=0.99\linewidth]{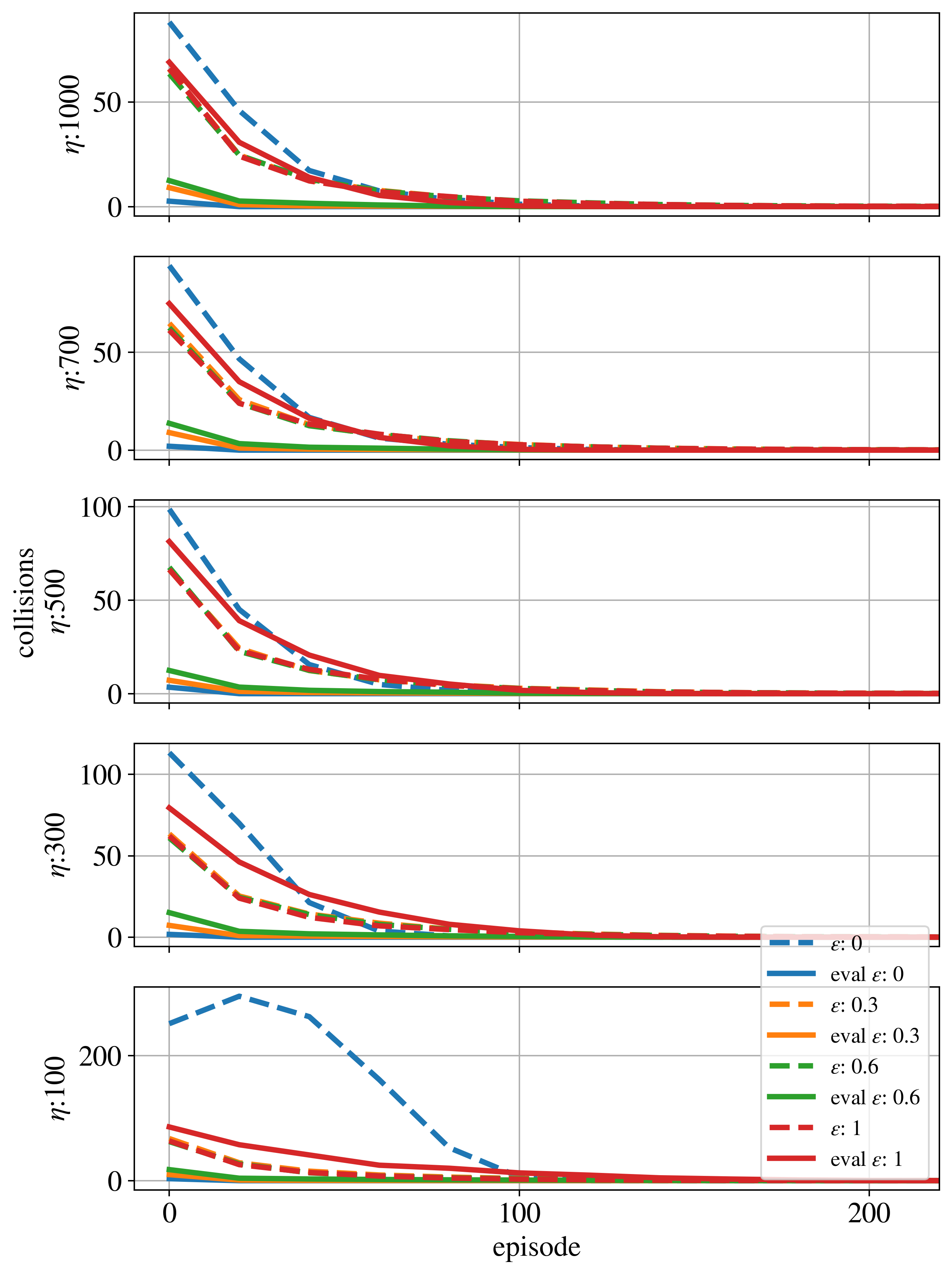}
    \caption{The collision convergence}
    \label{fig:kl_soften_d}
\end{subfigure}

\caption{Learning and evaluation curves of step length and collision rate under different KL regularization strengths and softening levels. Results are averaged over five independent runs. Softening the companion priors ($\epsilon=0.3,0.6$) consistently improves the efficiency–safety trade-off, particularly under stronger KL coupling (smaller $|\eta|$).}
\label{kl_soften_curves}
\end{figure}

Figure~\ref{kl_soften_curves} reports averaged learning and evaluation curves over multiple runs for step length and collision rate under different KL strengths $\eta$ and softening levels $\epsilon$. Across all settings, softened klDMP consistently achieves a better balance between efficiency and safety. In particular, intermediate values such as $\epsilon \in [0.3, 0.6]$ provide faster convergence in step length while maintaining low collision rates. When KL regularization is strong (small $\lvert \eta \rvert$), softening becomes especially beneficial, indicating that it stabilizes learning under aggressive coupling.

These results suggest that companion-prior softening serves as an effective stabilization mechanism for KCPR. By reducing the influence of immature companion priors during early learning, the proposed criterion stabilizes learning while preserving the long-term benefits of policy coupling. The proposed criterion better balances exploration, safety, and goal-directed efficiency than both pure klMP and fully entropy-regularized baselines.


\subsubsection{Effects of separate replay buffer}

We evaluate the effect of replay-buffer design by comparing MP and klMP-$\epsilon$ under three replay configurations: no replay buffer, a single shared replay buffer, and separate replay buffers for positive and negative experiences.
Experiments are conducted in a 36×19 three-room grid-world environment containing both goal and punishment signals. The no-buffer setting performs online updates after each transition, while buffer-based variants use episodic experience collection with mini-batch updates.

Figure~\ref{fig:sep_main} reports the learning curves across all methods and KL settings. Introducing any form of replay buffer significantly improves stability compared to the no-buffer baseline. More importantly, separating experiences by motivational source consistently accelerates convergence and improves both step efficiency and collision performance. A critical observation emerges under strong KL regularization (small $|\eta|$), where the coupling between reward-seeking and punishment-related learning becomes highly asymmetric. In this regime, both no-buffer and single-buffer variants fail to maintain stable navigation behavior, often collapsing into shortcut-dominated trajectories. In contrast, the separate-buffer design preserves informative negative transitions, enabling stable propagation of $Q^-$ and restoring balanced exploration between goal-seeking and risk-avoidance behaviors.

\begin{figure}[H]
  \centering
  \begin{subfigure}{0.7\linewidth}
      \centering
      \includegraphics[width=1\linewidth]{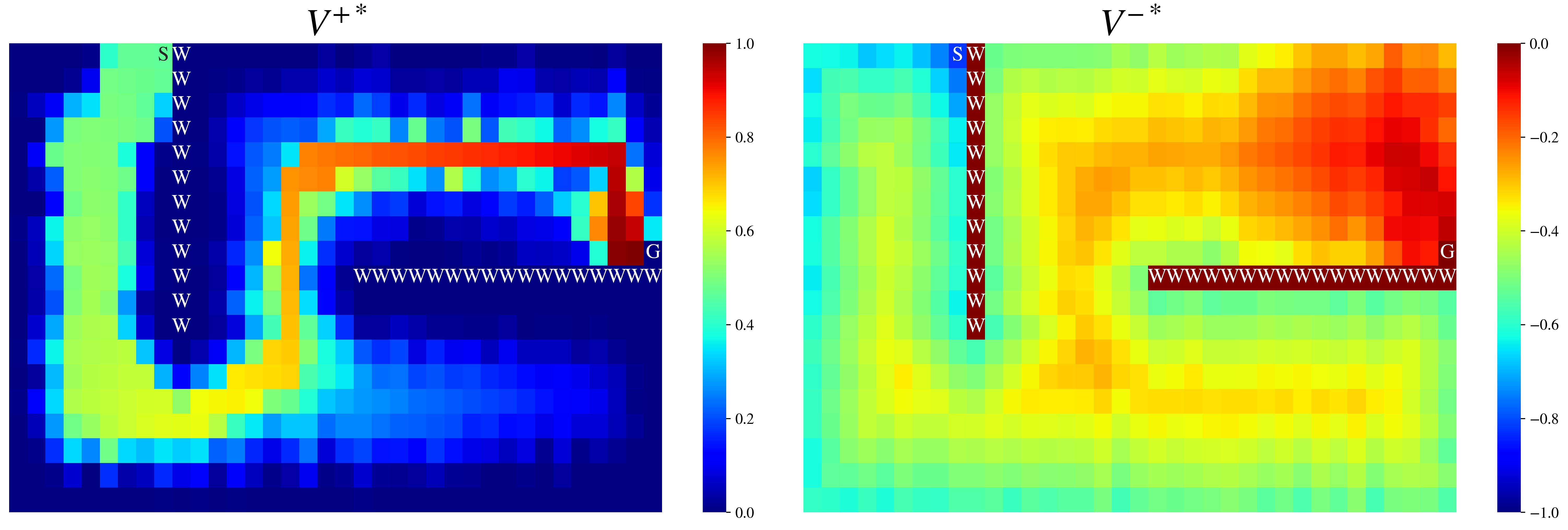}
      \caption{klMP with one buffer when $\epsilon=0.3$, $\eta=1000$}
      \label{fig:kl_one}
  \end{subfigure}
  
  \begin{subfigure}{0.7\linewidth}
      \centering
      \includegraphics[width=1\linewidth]{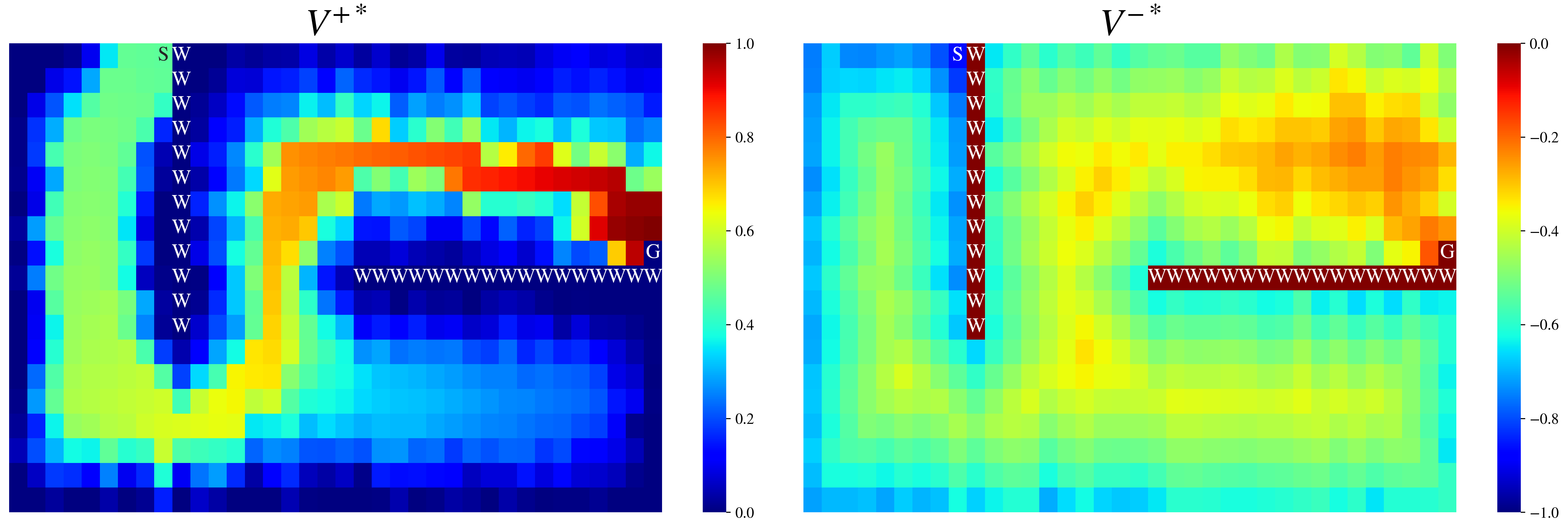}
      \caption{klMP with separate buffer when $\epsilon=0.3$, $\eta=1000$}
      \label{fig:kl_sep}
  \end{subfigure}
  
  \caption{Heatmaps of the learned reward-seeking and punishment-related state-value functions in the 36×19 Three-room maze using a single shared replay buffer and separate replay buffers. The separate replay-buffer design yields a more structured punishment-value landscape while preserving the reward-value function, leading to improved risk-aware learning.}
  \label{fig:kl_buf}
  \end{figure}

Figure~\ref{fig:kl_buf} further illustrates the effect of replay-buffer design on value propagation. While the learned reward-value function $V^+$ remains largely unchanged, separate replay buffers produce a substantially more structured punishment-value landscape than a single shared replay buffer. This indicates that separating positive and negative experiences facilitates the propagation of punishment-related value information without degrading reward learning.

These results show that disentangled experience replay substantially improves learning under reward–punishment decomposition. By reducing interference between reward- and punishment-related experiences, the separate-buffer design strengthens the propagation of punishment-related values and yields more stable learning.

\begin{figure}[H]
  \centering
  \setcounter{subfigure}{0}%
  \subcaptionbox{MP\label{fig:sep_a}}%
    {\includegraphics[width=0.8\linewidth]{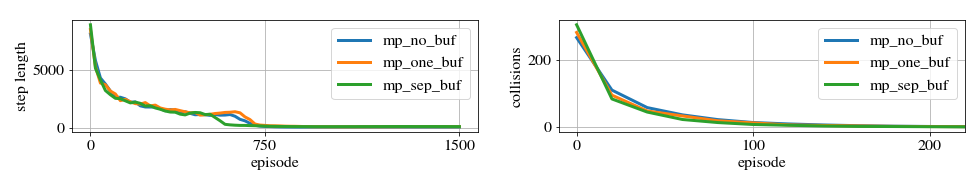}}
\end{figure}

\begin{figure}[H]
  \ContinuedFloat
  \centering
  \setcounter{subfigure}{1}%
  \subcaptionbox{klMP-$\epsilon$ when $\epsilon=0.3$\label{fig:sep_b}}%
    {\includegraphics[width=0.8\linewidth]{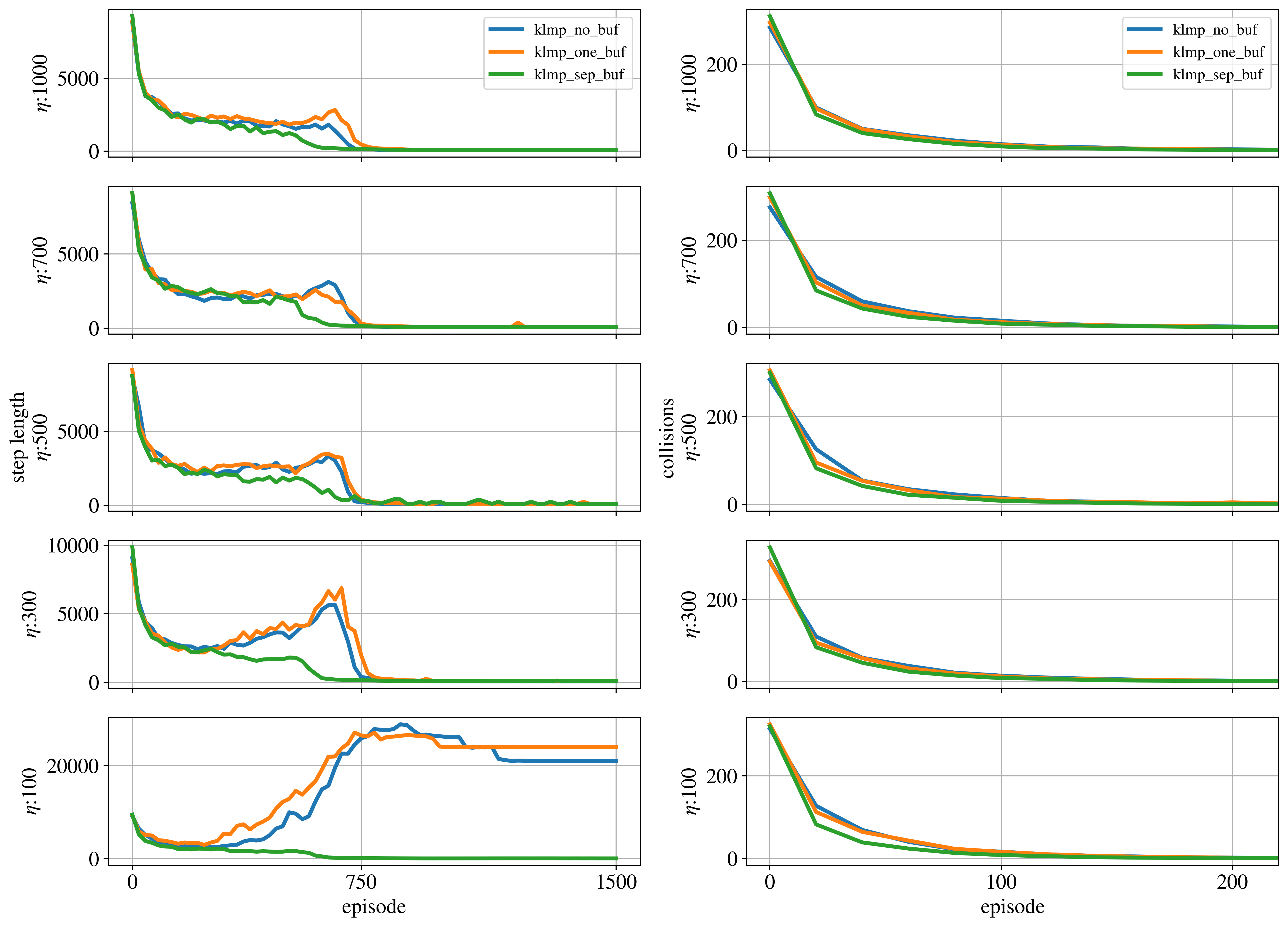}}
  \caption{Learning curves of step length and collision rate for MP and klMP-$\epsilon$ under three replay-buffer configurations (no buffer, single buffer, and separate buffers). The separate replay-buffer design consistently improves learning stability and convergence, with the largest benefit observed under stronger KL coupling (smaller $|\eta|$).}
  \label{fig:sep_main}
\end{figure}

\subsection{Gazebo navigation}

\begin{figure}[H]
    \begin{center}
        {{\includegraphics[width=0.8\hsize]{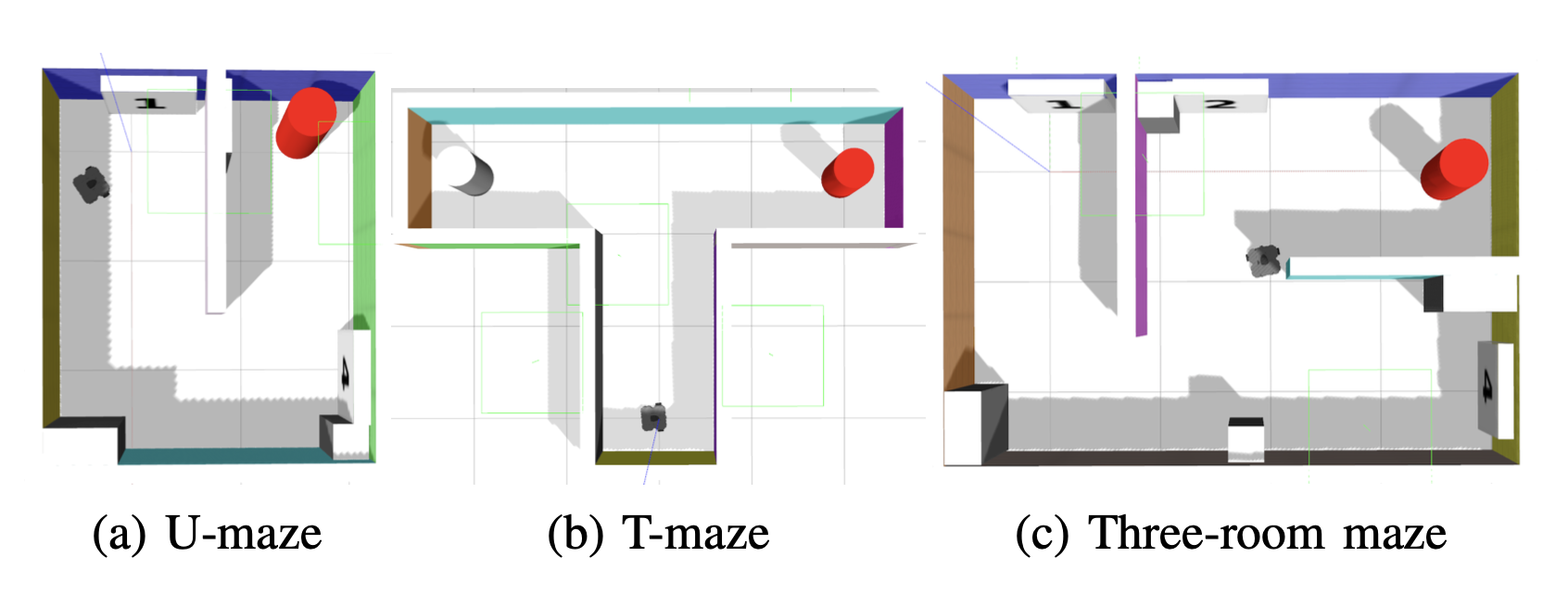} }}%
    \end{center}
    \caption{Three mazes with different complexity in Gazebo}
    \label{fig:mazes}
\end{figure}

We evaluate klDMP in simulated robotic navigation tasks using TurtleBot3 in ROS Gazebo. The objective is to investigate whether the proposed KL-coupled learning framework generalizes to continuous, noisy, and visually grounded navigation environments. Three maze environments of increasing complexity are considered: a U-maze, a T-maze, and a Three-room maze (Figure~\ref{fig:mazes}), which progressively increase the difficulty of long-horizon navigation, obstacle density, and ambiguity in safe path selection. In all environments, the robot is initialized with random orientations from a fixed start region and selects discrete motion primitives to navigate toward a goal cylinder (absorbing state). A sparse reward of $+5$ is provided upon reaching the goal, while collisions incur a penalty of $-0.5$. Observations consist of RGB images and LiDAR scans. All methods employ the same multimodal perception backbone for feature extraction to ensure a fair comparison. DQN, SQL, and softDMP estimate only action values on top of the shared representation, whereas klDMP additionally introduces dedicated value and policy heads for both the reward-seeking and punishment-related subsystems, implementing the KCPR/KCSO framework described in Section~\ref{sec:deep_realization}. Table~\ref{tab:comparison} summarizes the main characteristics of the compared methods.

The baseline methods are evaluated using their best-performing configurations reported in previous studies \citep{Wang2024}. Specifically, SQL adopts the $\eta=0$ ("mean") operator, while softDMP uses the best-performing regularization coefficient reported in the original work ($\eta=10000$). For klDMP, both the KL regularization coefficient and the companion-prior softening coefficient are selected through parameter scanning in the present work, yielding the best overall performance at $\eta=1000$ and $\epsilon=0.6$, respectively. Furthermore, since the Grid-world experiments demonstrated the superiority of the separate replay-buffer design for reward--punishment learning, all DMP-based methods employ separate replay buffers throughout the Gazebo experiments. For both softDMP and klDMP, behavior selection is generated by hard-max mixing of the reward-seeking and pain-avoiding value functions following \citep{Wang2021a}. All remaining hyperparameters are kept identical across methods.


\begin{figure}[htbp]
  \centering
  \begin{subfigure}[t]{0.48\linewidth}
      \centering
      \includegraphics[width=1\linewidth]{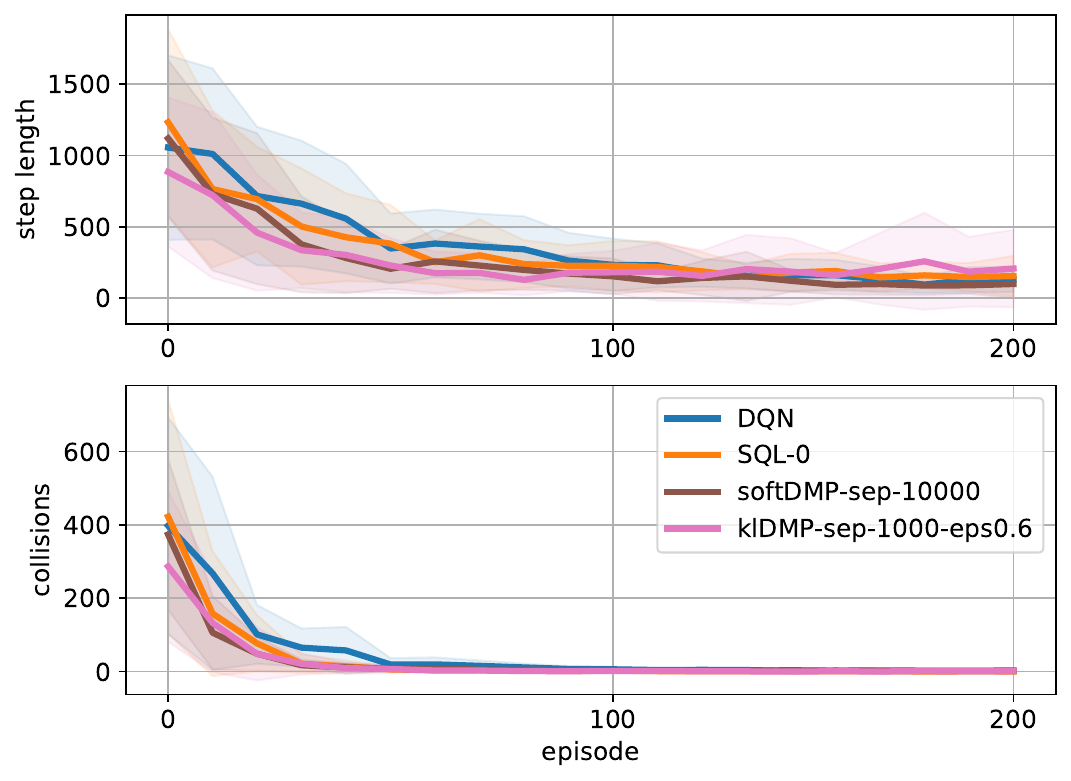}
      \caption{T-maze learning curves}
      \label{fig:t_curve}
  \end{subfigure}
  \hfill
  \begin{subfigure}[t]{0.465\linewidth}
      \centering
      \includegraphics[width=1\linewidth]{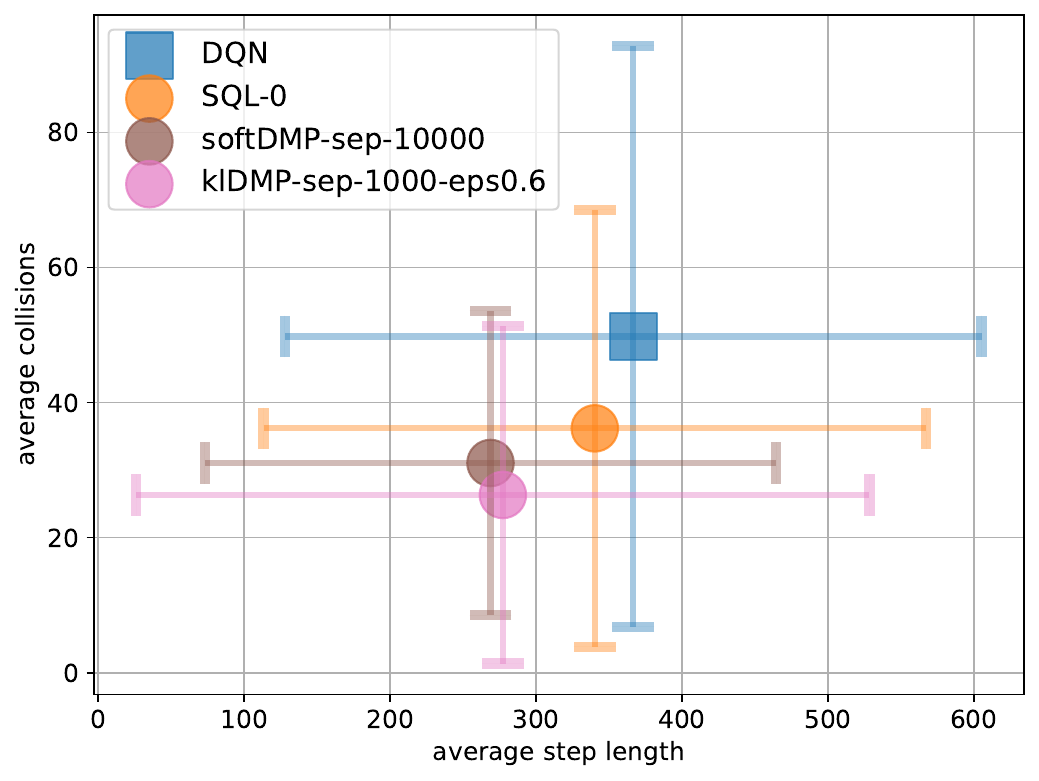}
      \caption{T-maze scattered metrics}
      \label{fig:t_scatter}
  \end{subfigure}
  \hfill
  \begin{subfigure}[t]{0.465\linewidth}
      \centering
      \includegraphics[width=1\linewidth]{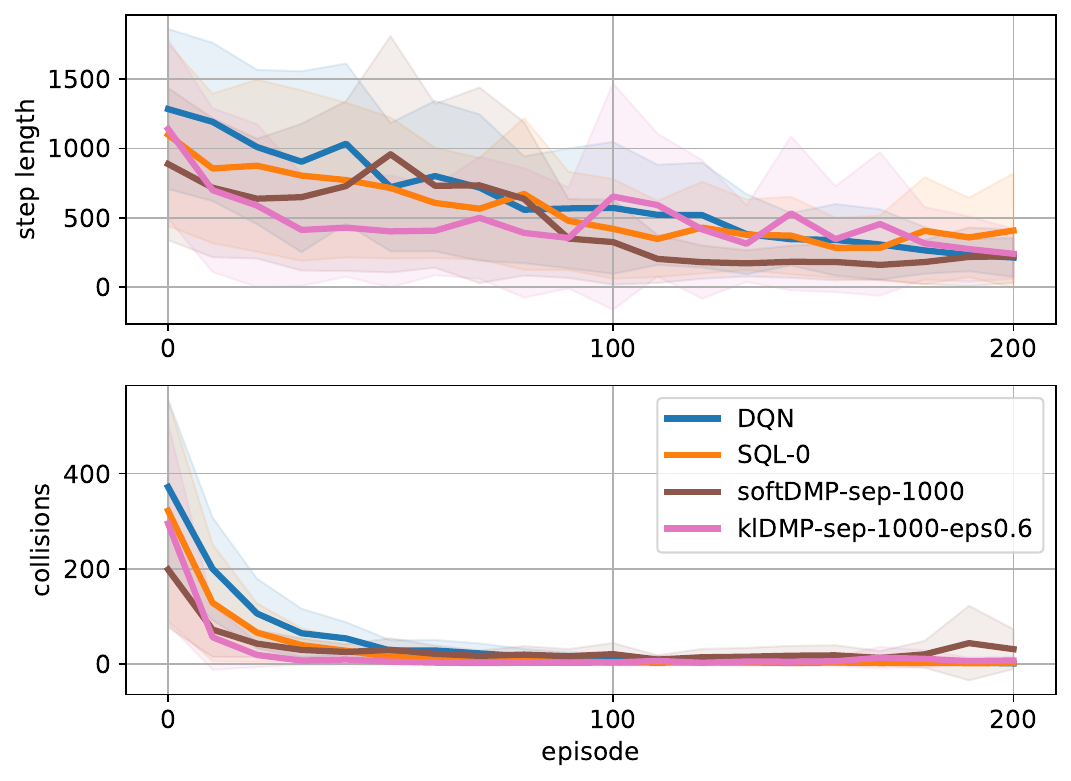}
      \caption{U-maze scattered metrics}
      \label{fig:u_curve}
  \end{subfigure}
  \hfill
  \begin{subfigure}[t]{0.465\linewidth}
      \centering
      \includegraphics[width=1\linewidth]{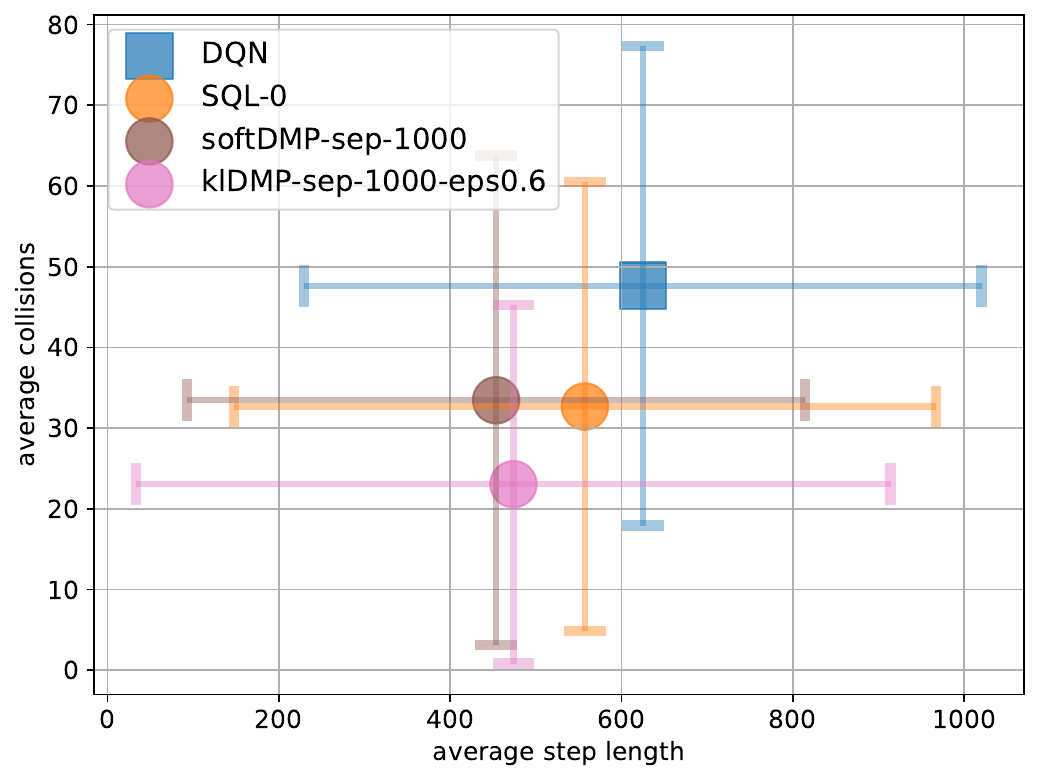}
      \caption{U-maze scattered metrics}
      \label{fig:u_scatter}
  \end{subfigure}
  \hfill
  \begin{subfigure}[t]{0.465\linewidth}
      \centering
      \includegraphics[width=1\linewidth]{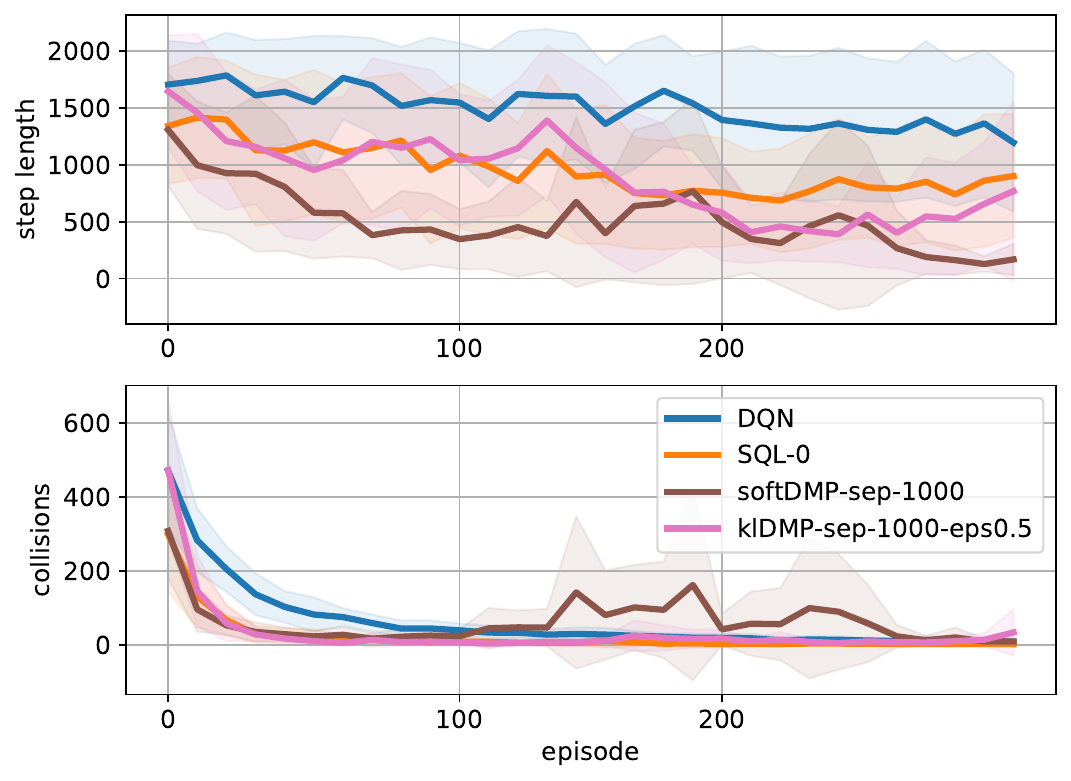}
      \caption{3R-maze scattered metrics}
      \label{fig:3r_curve}
  \end{subfigure}
  \hfill
  \begin{subfigure}[t]{0.465\linewidth}
      \centering
      \includegraphics[width=1\linewidth]{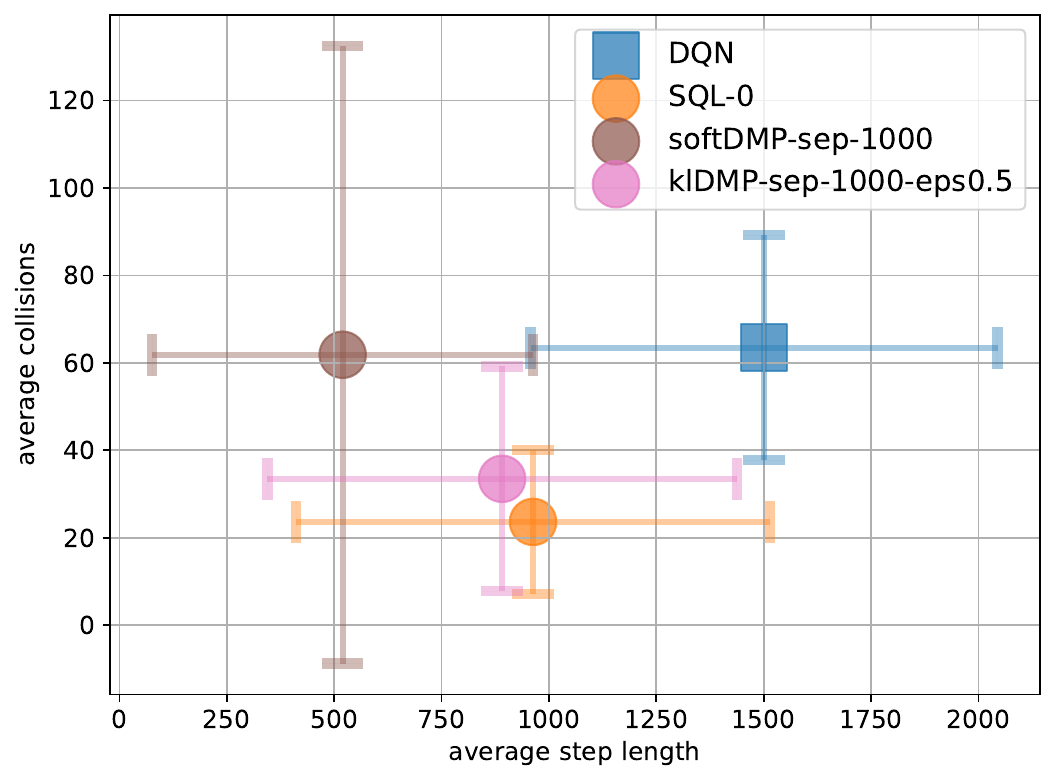}
      \caption{3R-maze scattered metrics}
      \label{fig:3r_scatter}
  \end{subfigure}
  
  \caption{Navigation performance of DQN, SQL, softDMP, and klDMP across the T-maze, U-maze, and Three-room Gazebo environments. The left column shows the learning curves of trajectory length and collision rate, illustrating the different learning dynamics of the compared methods. The right column summarizes the final trade-off between navigation efficiency and safety using the average trajectory length and collision rate over evaluation episodes. All results are averaged over five independent runs.}
  \label{fig:gazebo_res}
  \end{figure}

\begin{table}[t]
\centering
\caption{Comparison of reinforcement learning frameworks.}
\label{tab:comparison}
\footnotesize
\begin{tabular}{lccccc}
\hline
Method & Reward Signals & Prior Policy & Motivational Interaction & Operator \\
\hline
DQN
& Single
& --
& None
& Max \\

SQL
& Single
& Uniform
& None
& Soft \\

DMP
& Reward + Punishment
& --
& Value mixing
& Max/Min \\

softDMP
& Reward + Punishment
& Uniform
& Value mixing
& Soft \\

\textbf{klDMP}
& Reward + Punishment
& Companion
& \textbf{KCPR}
& \textbf{KCSO} \\
\hline
\end{tabular}
\end{table}

Figure~\ref{fig:gazebo_res} summarizes the navigation performance of all methods across the three environments. Although all methods successfully learn goal-reaching behaviors, they exhibit fundamentally different trade-offs between navigation efficiency and safety.

DQN produces the longest trajectories and the highest collision rates, particularly in the Three-room maze. Introducing entropy regularization through SQL substantially improves safety. As reported in our previous study \citep{Wang2024}, the SQL-$\eta=0$ ("mean") operator avoids the extreme behaviors induced by conventional max and min operators, yielding more balanced action evaluation and consistently lower collision rates. However, because SQL optimizes only a single reward function, its safety emerges from entropy-based averaging rather than an explicit representation of risk, resulting in conservative navigation and longer trajectories.

softDMP further improves early learning by decomposing reward and punishment into separate value functions. One possible explanation is that the punishment-related subsystem initially exerts a stronger influence on the shared behavior policy, resulting in rapid collision reduction during early training. However, because the two subsystems remain independently optimized and interact only through the behavior policy, safety knowledge cannot directly influence reward-value learning. This limitation is particularly evident in the Three-room maze, where efficient and safe behaviors are more difficult to reconcile, leading to higher collision variability despite efficient trajectories.

In contrast, klDMP consistently maintains competitive navigation efficiency while producing more stable collision performance than softDMP. Although convergence is slightly slower during early training, companion-policy regularization and KL-coupled Bellman updates enable the reward-seeking and punishment-related subsystems to interact directly during optimization. One possible explanation is that these optimization-level interactions stabilize value propagation as the companion policies become increasingly informative during training, leading to more consistent policy learning and, ultimately, more stable navigation performance.

These results suggest that the primary benefit of KL coupling is not faster optimization but more stable coordination between reward-seeking and punishment-related learning. Compared with softDMP, which combines independently optimized subsystems only through the behavior policy, klDMP embeds the interaction directly into the optimization process, resulting in more stable navigation while maintaining competitive efficiency.

\section{Conclusions}

This paper introduced KL-Coupled Policy Regularization (KCPR), a policy coordination framework for RPRL, together with its practical realization, klDMP. By treating companion sub-policies as dynamically learned priors, KCPR enables direct interactions between reward-seeking and punishment-related learning processes. We further derived KL-Coupled Soft Optimality (KCSO), which yields coupled soft-optimal policies and KL-regularized Bellman operators. Experiments in grid-world and Gazebo navigation tasks demonstrated that KL coupling improves safety and learning stability while maintaining competitive task performance.

Beyond its algorithmic contribution, KCPR provides a new perspective on coordinating multiple decision policies through learned behavioral priors. Rather than optimizing reward and punishment independently, the proposed framework allows interacting policies to continuously shape one another's optimization process. More broadly, these results suggest that policy coordination may serve as a useful design principle for reinforcement learning systems with multiple behavioral objectives.

Several directions remain for future work. In the current framework, both the KL regularization strength and the softening parameter are manually specified. Developing adaptive mechanisms for regulating policy coupling and exploration may further improve robustness across tasks. Although this work focuses on discrete-action navigation, extending the proposed framework to continuous-control reinforcement learning remains an important direction for future work. More broadly, the same framework could naturally be extended to additional motivational systems, such as curiosity, uncertainty, and homeostatic regulation.

\bibliography{main}
\bibliographystyle{elsarticle-num-names}
\end{document}